# Utilizing Cognitive Signals Generated during Human Reading to Enhance Keyphrase Extraction from Microblogs

Xinyi Yan[1], Yingyi Zhang[2], Chengzhi Zhang[1, *]

1. Department of Information Management, Nanjing University of Science and Technology, Nanjing, 210094, China
2. Department of Archives and E-government, Soochow University, Suzhou, China

**Abstract**: Microblogging platforms have seen exponential growth, leading to an abundance of user-generated content. The challenge now is to efficiently extract crucial information from this vast and dispersed text data. It also serves as the goal of our research on Automatic Keyphrase Extraction (AKE) for Microblog. Eye-tracking signals, that reflect users' tendency to prioritize certain words while reading, have been employed to enhance AKE performance from Microblogs. However, relying solely on eye-tracking has its limitations owing to constraints in physiological mechanism support, acquisition techniques, and feature decoding. Consequently, we propose the integration of electroencephalogram (EEG) signals with eye-tracking signals to improve Microblogs-based AKE, thereby overcoming the aforementioned limitations. Our first step is identifying specific features present in cognitive signals generated during human reading. We selected EEG signals (8 features) and eye-tracking signals (17 features) from the cognitive language processing corpus ZUCO, to examine the efficacy when they are combined with the Microblogs-based AKE. To avoid cognitive signal distortion by certain model structures, we introduced these signals at the inputs of the soft attention layer and at the query vectors of the self-attention layer. For evaluation, we performed several AKE tests on Microblogs with various combinations of cognitive signals. The results demonstrate a consistent enhancement in the performance of AKE due to cognitive signals generated during human reading, regardless of different feature combinations and models. Specifically, EEG signals exhibited the most significant improvement. However, combining EEG signals with eye-tracking signals yielded results that fell between the performance levels of the two signal types, indicating that their integration might have some synergistic effects. Further investigation is needed to understand the underlying mechanisms responsible for this outcome. The code and dataset for this paper can be accessed at https://github.com/yan-xinyi/AKE.

---

* Corresponding author, Email: zhangcz@njust.edu.cn

## 1. Introduction

With the rapid development of Microblog platforms such as Twitter, Facebook, and Instagram, user-generated content has exhibited an explosive growth trend. How to better extract crucial information from this vast and unstructured data has become a common concern for researchers and businesses. Due to the overwhelming amount of unorganized Microblog posts, many users struggle to find their desired content. The emergence of Hashtags has resolved this dilemma by automatically categorizing Microblogs into message sets based on the same topic, which facilitating user to search and browse. Hashtag plays an essential role in the analysis of massive and unstructured data. It can help researchers and enterprises uncover new techniques and intelligence from massive amounts of blog posts, which in turn aids their research and decision-making. However, fewer than 15% of Microblogs contain Hashtags (Wang et al.,2011), making it challenging for users to discover desired information from posts lacking tags. To cope with this situation, many studies have focused on Hashtag extraction and analysis (Li et al., 2016; Zheng & Sun, 2019).

Automatic Keyphrase Extraction (AKE) is a crucial task in Natural Language Processing (NLP). Its objective is to extract words or phrases that represent the topic of the given text (Turney, 2002). AKE is commonly used in Hashtag research, and its principle is similar to that of Hashtag extraction. However, many Microblogs lack Hashtags currently, leading to aggregated and redundant posts outside the retrieval network. Thus, utilizing AKE to automatically assign Hashtags to Microblogs is vital. For instance, in product market research, merchants can extract Hashtags to discern the needs and intentions of potential users (Jeong et al., 2022). Furthermore, understanding the interest preferences of specific users through their browsing, following, and liking behaviors allows for more accurate personalized recommendation services (Kumar & Srinathan, 2008). Additionally, AKE from Microblogs is also used for opinion mining (Messaoudi et al., 2022),sentiment analysis (Vashishtha & Susan, 2021), user engagement analysis (Gkikas et al.,2022) and emergency response (Dutt et al., 2018) and other scenarios..

Reading is commonplace, often seen as a natural behavior involving thinking, feeling and imagining. However, the reality is far more complex; underlying these activities are cognitive processes that play a crucial role in human development. These cognitive processes trigger the release of chemicals from neurons in the brain, generating cognitive signals composed of both chemical and electrical components. These signals, produced during reading behaviors, provide valuable insights into our understanding of the human language system. They have proven to be a rich resource for NLP tasks (Hollenstein et al., 2019). Advances in signal acquisition and processing techniques have lowered the cost of acquiring cognitive signals. This led to the growing popularity of research using cognitive signals to improve various NLP tasks. Among these signals, eye-tracking data stands out as it provides essential insights into a reader's attention distribution and is relatively easy to acquire. The use of eye-tracking signals in NLP research has matured, finding applications in areas like lexical annotation (Barrett et al., 2016), semantic analysis (Porzio et al., 2021), and sentiment recognition (Tarnowski et al., 2020).

Nonetheless, there are still certain limitations when it comes to using single-source cognitive signals, particularly eye-tracking signals, in NLP research. Firstly, the basis for inferring cognitive activity occurring in the brain from eye-tracking signals is relatively weak. When analyzing a set of eye-tracking data, researchers usually aggregate the effects of various concurrent cognitive activities. However, direct inference of cognitive activities from eye-tracking recordings is not feasible (Kok & Jarodzka, 2017). Secondly, the study of human readings mechanisms using eye-tracking signals necessitates devices with high sampling rates and accuracy (Hollenstein et al., 2020). While eye-tracking plug-ins based on low-cost devices like Webgazer (Papoutsaki et al., 2016) have become popular, they tend to have significantly higher data omission rates (Funke et al., 2016). Due to limitations in physiological mechanisms and data acquisition techniques, eye-tracking signals do not exhibit outstanding sensitivity and precision when compared to other existing cognitive signals (Bueno et al., 2019). Thirdly, among the existing studies, study by Zhang & Zhang (2021) is the most representative. They conducted qualitative and quantitative experiments, confirming that eye-tracking signals have a positive impact on Microblog-based AKE. However, the observed improvement is not substantial, as the enhancement level is only around 1%. Additionally, this study falls short in terms of model selection and feature considerations. From a model perspective, the models constructed in this research are limited to Bi-LSTM and attention-based improved models. From a feature perspective, the eye-tracking signals only consider the feature of human readings time, resulting in insufficient feature quantity and granularity.

In order to address the limitations mentioned above, in this paper, we incorporates both electroencephalogram (EEG) signals and eye-tracking signals into the Microblog-based AKE research. Specifically, we propose corresponding solutions: First, EEG signals directly reflect the electrical activity of brain neurons, offering comprehensive cognitive information regarding human readings processing. Research by Scharinger et al. (2020) suggests that EEG signals are more closely related to potential brain activities compared to eye-tracking signals. Our study confirms the feasibility of incorporating EEG signals into single eye-tracking research. Physiologically, derived indices of EEG signals, like regional indices, can monitor the activation state of specific functional brain regions (Zhu et al., 2021), allowing for more accurate inference of specific cognitive activities. Second, cognitive signals generated during human reading exhibit a complementary relationship (Hollenstein et al., 2020). Combining neurophysiological markers provided by EEG signals with behavioral information recorded by eye-tracking signals enables a more comprehensive understanding of brain perception, attention and cognitive processes during natural reading (Plöchl et al., 2012). Previous research has demonstrated that the combination of these signals outperforms single eye-tracking mode in tasks like vigilance estimation(Zheng & Lu, 2017), emotion recognition and information extraction (Hollenstein et al., 2018). To address the limitations in single eye-tracking mode data collection, electrooculography (EOG) can be used to enhance and calibrate the accuracy of eye-tracking signals by combining it with EEG signals (Sheoran & Saini, 2020). Finally, to evaluate the effect of integrating cognitive signals as external features on the performance of Automated Keyphrase Extraction (AKE) from Microblogs, conducting controlled experiments is crucial. These should involve varying sources of cognitive signals, different feature combinations within the same cognitive signal source, and diverse combinations of features across different cognitive signal sources.

Our contributions are the following four folds:

(1) In this paper, we applied different types of cognitive signals generated during human reading to AKE from Microblogs for the first time. Specifically, we combine EEG signals and Eye-tracking signals jointly to AKE based on the open-source cognitive language processing corpus ZUCO. Moreover, we successfully verified the enhancement of EEG signals and Eye-tracking signals on six different structured models: BiLSTM, BiLSTM+CRF, ATT-BiLSTM, ATT-BiLSTM+CRF, SATT-BiLSTM and ATT-BiLSTM+CRF.

(2) We compared the effects of different frequency bands of EEG signals on the performance of the AKE. The results indicated that EEG signals in the β and γ frequency bands demonstrate the most significant enhancement in AKE from Microblogs. Based on existing research, these two frequency bands are closely related to the cognitive workload during reading, and there exists a mutually constraining relationship between two of them.

(3) Furthermore, we evaluated AKE by combining the most effective EEG signals and eye-tracking signals from single-source cognitive signal tests, but the model's performance did not exhibit any further improvement.

(4) Analyzing the unsatisfactory results of the previous experiments, we improved the AKE model based on Pre-trained Language Models (PLMs): First, we incorporate Glove embeddings into the input layer of the SATT-BiLSTM+CRF model, which exhibited the best AKE test performance. Second, we propose an improved AKE based on BERT. Lastly, we implemented an improved AKE based on the T5 (including T5-Base and T5-Large). According to the experimental results, it shows that T5-Large model can maximize the performance of the AKE without weakening the weights of cognitive features.

All the data and source code of this paper are freely available at the GitHub website: https://github.com/yan-xinyi/AKE.

## 2. Related Work

Our study focuses on improving the performance of AKE from Microblogs by leveraging EEG signals and Eye-tracking signals jointly. Therefore, we provide an overview of the research on AKE from Microblogs in Section 2.1. In Section 2.2, we summarize the existing studies on the application of EEG signals and Eye-tracking signals in the field of NLP.

### 2.1 Microblogs Automatic Keyphrase Extraction

Microblog is a form of online social media, which usually refers to short texts of less than 140 words posted on social media platforms such as Twitter, Facebook, Instagram (Gaonkar et al.,2008). Hashtag, denoted by the "#" symbol, is commonly employed to mark and organize important information or topics within Microblogs. In the realm of social media text analysis, Hashtag holds significant value. They enable users to swiftly search and monitor specific topics, facilitating information dissemination and fostering interactions. Moreover, Hashtag serves as a tool for website administrators to curate and manage information on social media platforms(Potnis & Tahamtan, 2021). Nevertheless, a mere 15% of the existing Microblogs include Hashtags. To address the challenge posed by the vast volume of unstructured Microblogs, automatic Hashtag extraction has garnered substantial research attention in recent years(Gong et al., 2018). This research focus has also paved the way for various downstream applications, including user research, analysis of public opinion(Zahera et al.,2021) and event tracking(Sakaki et al., 2010; Zheng & Sun, 2019). AKE shares a fundamental similarity with Hashtag extraction and is widely employed in Hashtag research. Many scholars have contributed to comprehensive AKE reviews(Nasar et al., 2019; Xie et al., 2023), which establishing the foundation for subsequent research in methods, data, evaluation metrics and other aspects. The methods commonly utilized for AKE from Microblogs can be categorized into unsupervised and supervised approaches.

**(1) Unsupervised methods**

Unsupervised AKE eliminate the need for manual corpus annotation. Instead, this method relies on statistical features, word graph, topics, embeddings and other Microblog text information to discriminate candidate keyphrases. As Table 1 shown, we categorize unsupervised AKE into 4 classes.

| Categories of methods | （Authors, Year of publication) | Methods proposed | Main findings |
|---|---|---|---|
| Statistical-base methods | *El-Beltagy & Rafea, 2009* | *KPMiner* | This system does not need to be trained on a particular document set, and is capable of extracting keyphrases from both English and Arabic documents. |
| | *Campos et al., 2018* | *YAKE!* | This method integrates the positional information of keywords, and employs innovative statistical metrics to capture the contextual information of candidate words. |
| | *Won et al., 2019* | *KCRank* | A statistical method for extracting keyphrases at document level, combining simple heuristic rules, which can compete with state-of-the-art systems. . |
| | *Kang et al., 2021* | *SSDIPA* | A semantic similarity algorithm based on WordNet semantic dictionary toward distance, information and property (SSDIPA) is proposed, which quantifies the degree of association between words. |
| | *Hassani et al., 2022* | *LVTIA* | Lecture Video Text mining-base Indexing Algorithm (LVTIA), a new method for video lecture indexing using statistical features. |
| Graph-based methods | *Mihalcea & Tarau, 2004* | *TextRank* | As a graph-based ranking model for text processing, TextRank does not require deep linguistic knowledge, nor domain or language specific annotated corpora, which makes it highly portable to other domains, genres, or languages |
| | *Bellaachia & Al-Dhelaan, 2012* | *NE Rank* | A novel unsupervised graph-based keyword ranking method. Several experiments have shown the potential of NE-Rank approaches with 16% to 39% improvement. |
| | *Bougouin et al., 2013* | *TopicRank* | Results show that TopicRank significantly outperforms state-of-the-art methods on *SemEval, WikiNews* and *DEFT* datasets. |
| Topic-based methods | *Song et al., 2009* | - | This paper presents several topic and keyword re-ranking approaches that can help users better understand and consume the LDA-derived topics in their text analysis. |
| | *Siu et al., 2014* | *SOU* | SOU method together reduce the topic verification equal error rate from 12% to 7%, Some selected SOU n-grams are highly correlated with the keywords. |
| | *Chin et al., 2019* | *TOTEM* | TOTEM leverages LDA to generates the topic labels as well as the topic summaries associated with them. Evaluation of the generated topic summaries suggest reasonably high consistency and effectiveness of summaries. |
| | *Rodrigues et al., 2021* | - | The LDA technique for trend analysis resulted in an accuracy of 74% and Jaccard with an accuracy of 83% for static data. |
| Embedding-based methods | *Bennani-Smires et al., 2018* | *EmbedRank* | A novel unsupervised method, that leverages sentence embeddings. |
| | *Sun et al., 2020* | *SIFRank* | A New Baseline for Unsupervised Keyphrase Extraction Based on Pre-Trained Language Model |
| | *Liang et al., 2021* | - | The results show that this model outperforms most models while generalizing better on input documents with different domains and length. |

**Table 1.** Summary of representative studies of unsupervised keyphrase extraction approaches

**Statistical-base methods**: Common statistical metrics include word weight metrics (word nature, word frequency, word length, etc.), word document location metrics and word association information metrics. In recent years, researchers have made efforts to enhance the performance of AKE through the integration of external knowledge, such as lexical knowledge bases (Jiang, 2008; Li & Zhao, 2016), external features (Campos et al., 2018; Goz & Mutlu, 2023; Kang et al., 2021) and external models (Koloski et al., 2022). Furthermore, multimodal information extraction has emerged as a prominent research area. Hassani et al.(2022) introduced a novel approach that combines video frame text and audio text to comprehensively consider local and global features of candidate phrases found within the audio signal or video frame.

**Graph-based methods**: It is to identify keyphrases within a document's linguistic network graph by analyzing its structure and features. TextRank (Mihalcea & Tarau, 2004) is a classic and efficient graph-based method, which has been further extended to include several improved variations such as ExpandRank (Wan & Xiao, 2008) and TopicRank (Bougouin et al., 2013). In the study conducted by Fushimi & Kanno (2020), a graph was constructed to connect similar microblogs, allowing for the extraction of user opinions and needs regarding specific objects or events from the resulting connected components. Graph-based methods, in comparison to statistical feature-based approaches, exhibit higher accuracy and versatility across various text data types. Notably, existing methods tend to focus solely on the importance of candidate key phrases while overlooking other critical factors, such as uninformative sentences. Vega-Oliveros et al. (2019) made a pioneering attempt in unsupervised keyphrase extraction by incorporating hierarchical multi-granularity features to leverage sentence saliency. They also introduced a novel federated model based on location-aware graphs, paving the way for advancements in this area.

**Topic-based methods**: This approach models text themes to extract relevant keyphrases. Despite being a relatively recent development, the utilization of topic modeling technique in AKE has gained substantial traction. Boudin (2018) introduced an unsupervised extraction model that incorporates keyphrase selection preferences into a multi-partite graph structure's candidate ranking mechanism to encode topic information. To address the challenge of unfeasible Twitter summarization on mobile devices, Chin et al.(2019) developed TOTEM, a lightweight, personalized, topic-based Twitter summarization engine designed for mobile devices.

**Embedding-based methods**: In recent years, embedding-based AKE research has gained popularity due to advances in representation learning. Early embedding methods(Bennani-Smires et al., 2018; Sun et al., 2020) focused on calculating the similarity between word embeddings and document embeddings, leading to potential confusion regarding word importance in different contexts. To address this, Liang et al.(2021) introduced an unsupervised AKE model that incorporates both local and global contextual information, effectively resolving context ambiguity.

**(2) Supervised methods**

Supervised methods, including Machine Learning (ML) based methods, Deep Learning (DL) based methods, Pre-train Language Models (PLMs) based methods and Large Language Models (LLMs) based methods, have been extensively used in Hashtag extraction from Microblogs. Table 2 lists representative researches of each category of supervised AKE methods.

| Categories of methods | Authors, Year of publication | Methods used | Main findings |
|---|---|---|---|
| ML-based methods | *Witten et al., 1999* | *KEA* | Results show that Kea can on average match between one and two of the five keyphrases chosen by the author in this collection. |
| | *Mizuka et al., 2017* | *SVMs* | This work extracting commentary tweets from commentary tweet candidates using SVM. The result shows good suitability of the proposed method for tweet text. |
| | *Wang et al., 2018* | *RF* | With the help of feature extraction, discretization, and classification, the prediction accuracy has been improved to a large extent. |
| | *Zhang, 2008* | *CRFs* | Experimental results show that the CRF-based AKE outperforms other machine learning methods such as SVMs, multiple linear regression model etc. |
| DL-based methods | *Wang et al., 2006* | *FNN* | This approach is competitive with other known methods and practical, especially in the situation where the keyphrases are unavailable. |
| | *Zhang & Zhang, 2019* | *RNN* | This work is the first to utilize human attention on AKE tasks. The proposed models have proved to bring significant improvements on two Twitter datasets. |
| | *Ray Chowdhury et al., 2019* | *LSTM* | The model's performance is improved on both general Twitter data and disaster-related Twitter data. |
| | *Wu et al., 2018* | *Attention based LSTM* | Experiments on real-world datasets show that the proposed model outperforms several automatic baselines as well as humans in this task. |
| PLM-based methods | *Xiong et al., 2019* | *BLING-KPE* | Experimental results on OpenKP confirm the effectiveness of BLINGKPE. Zero-shot evaluations on DUC2001 demonstrate the improved generalization ability of learning from the open domain data compared to a specific domain. |
| | *Wang et al., 2020* | *SMART-KPE-full* | A combination of effective strategy induction and strategy selection within this approach for the KPE task outperforms state-of-the-art models. |
| | *Song et al., 2021* | *KIEMP* | Experimental results on six benchmark datasets show that KIEMP outperforms the existing state-of-the-art keyphrase extraction approaches in most cases. |
| LLM-based methods | (Wei et al., 2023) | *ChatIE* | Empirical results show that ChatIE achieves impressive performance and even surpasses some full-shot models on several datasets (e.g., NYT11-HRL). |
| | (Hu et al., 2023) | - | The findings revealed that ChatGPT outperformed GPT-3 in the zero-shot setting, with F1 scores of 0.418 (vs.0.250) and 0.620 (vs. 0.480) for exact- and |
| | (Lee et al., 2023) | *Galactica* | The F1 score of proposed method was ten times better than that of previous studies, and 42.7% of the generated keywords are relevant to author-defined keywords. |

**Table 2.** Summary of representative studies of supervised keyphrase extraction approaches

**ML-based methods.** In comparison to unsupervised methods, machine learning-based approaches demonstrate superior adaptability to various corpora and consistently deliver favorable experimental outcomes. Consequently, there has been a significant emphasis in numerous studies on leveraging machine learning techniques for keyphrase extraction. The KEA Algorithm (Witten et al., 1999), is a classical machine learning approach rooted in a classification problem. Based on this, subsequent research has focused on optimizing features and algorithms. Various features have been considered, including lexicality, mutual information and word linkage. Regarding algorithms, methods such as Support Vector Machines (SVMs, Mizuka et al., 2017), Random Forest (RF, Wang et al., 2018), and Conditional Random Fields (CRFs, Zhang, 2008) have been utilized to the Micro-blogs based AKE.

**DL-based methods.** The groundbreaking study conducted by Wang et al. (2006) utilized feedforward neural networks to supervised AKE. As a result, deep learning-based approaches have gained significant traction and become a prominent research focus in the field of AKE. Subsequently, researchers have employed a series of methods, including Recurrent Neural Network (RNN, Zhang & Zhang, 2019), Long Short-Term Memory Networks (LSTMs, Ray Chowdhury et al., 2019), and Attention mechanisms (Wu et al., 2018) to achieve higher performance of supervised AKE. The presence of non-standard expressions, such as abbreviations, misspellings and colloquialisms in Microblog significantly degrades the quality of keyphrases.

**PLM-based methods.** PLMs has gradually become the mainstream of supervised AKE research. To better evaluate the candidate phrases, Xiong et al. (2019) proposed BLING-KPE which regards AKE as an n-gram level keyphrase chunking task. Specifically, BLING-KPE incorporates ELMo embeddings (Peters et al., 2018) into a convolutional transformer network to model n-gram representations. Patel & Caragea (2019) studied the sensitivity of CRFs based on word embedding types pre-trained on Google News as well as those trained on a large collection of ACM research papers. Song et al. (2023) presented a review of the recent studies based on pre-trained language models, suggesting that the AKE methods with PLMs embedding outperform approaches with static embeddings in most cases. The current state-of-the-art research in PLM-based AKE include methods such as SMART-KPE-full (Wang et al., 2020), RoBERTa-JointKPE , and KIEMP (Song et al., 2021).

**LLM-based methods.** The term "large language models" was coined to encompass models like GPT-3 and its subsequent iterations, aiming to differentiate them from smaller pre-trained language models (Zhao et al., 2023). These models are distinguished by their immense scale, incorporating billions of parameters that empower them to acquire intricate patterns from language data. The advent of LLMs such as T5 (Raffel et al., 2023), GPT-4 (OpenAI, 2023) and FLAN-T5 (Chung et al., 2022) has led to a shift in the pre-training approach for large models. Instead of focusing on specific NLP tasks, the emphasis of LLMs lies in guiding the models to cultivate multi-task reasoning capabilities through downstream fine-tuning (Bao & Zhang, 2023). Wei et al. (2023) introduced ChatIE, a multi-turn QA framework built upon ChatGPT, aimed at enhancing the zero-shot information extraction capability of the model. Lee et al.(2023) introduced an algorithm using the Meta-pretrained large-scale language model Galactica for generating keywords, resulting in a tenfold increase in $F_1$ score compared to previous studies.

### 2.2 Application of EEG signals and Eye-tracking signals in NLP

Electroencephalogram (EEG) is a measurement of changes in scalp surface potentials caused by the activity of billions of neurons inside the brain. Due to its property of high temporal resolution, ease of operation and non-invasiveness (Khosla et al., 2020), EEG is considered one of the most widely used potential non-invasive tools for human brain cognitive activities study. Compared to detection techniques such as MRI (Functional Magnetic Resonance Imaging) and PET (Positron Emission Tomography), EEG is relatively cost-effective while providing more relevant and abundant information on brain cognitive activities. Whether in neurophysiology and psycholinguistics research or clinical neuropsychological diagnosis, EEG is considered a highly valuable detection technique (Gavaret et al., 2023).

**Study of EEG signals based NLP.** Electroencephalogram (EEG) signals are frequently used for NLP tasks such as machine translation, pattern recognition, language perception and other text processing tasks. Machine translation aims to computationally translate natural languages, involving aspects of natural language understanding, generation and translation. Ganushchak et al.(2011) learned the brain's response in different language environments through establishing a higher-performance language translation model with EEG signals. Pattern recognition tasks include emotion recognition, fatigue recognition, etc. Liu et al.(2023) proposed an EEG emotion recognition algorithm based on a global to local feature aggregation network (GLFANet). This study addressed the challenge of effectively improving emotion recognition task performance from the feature dimension. Peng et al.(2023) proposed a multi-feature fusion network (TA-MFFNet) based on time domain network and attention network. The results proved that the model can improve fatigue state recognition by learning valuable information from EEG signals in multiple dimensions.

| Categories of methods | Authors, Year of publication | Specific Task | Main findings |
|---|---|---|---|
| | *Ganushchak et al., 2011* | *Machine translation* | Overt speech production can be successfully studied using electrophysiological measures, for instance, event-related brain potentials (ERPs). |
| | *De Vito et al., 2018* | *cognitive ability assessment* | Ignoring working-memory representations has affective consequences, the contribution of selective-attention mechanisms to a wide range of human thoughts and behaviors leads to devaluation. |
| | *Scaltritti et al., 2020* | *language perception* | The comparison between tasks thus suggests that different beta desynchronizations reflect distinct EEG landmarks for language and motor processing |
| Study of EEG signals based NLP | *León Rodríguez et al., 2022* | *cross-linguistic studies* | They predicted that lower L2-exposure should produce less efficient L2 word recognition processing at the behavioral level, alongside neurophysiological changes at the early pre-lexical and lexical levels, but not at a post-lexical level |
| | *Liu et al., 2023* | *Emotion recognition* | The experiment results show that the proposed algorithm achieves higher accuracy on DEAP, SEED and DREAMER contrasted to other advanced algorithms. |
| Study of Eye-tracking signals based NLP | *Geraets et al., 2021* | *Emotion recognition* | Findings support the utility of VR emotional stimuli for assessment and training. |
| | *Jin et al., 2023* | *online comment analysis* | Two eye-tracking studies confirm these results from a physiological standpoint and reveal the attentional allocation during information processing. |
| | *Zhang & Zhang, 2021* | *Keyphrase extraction* | This article aims to leverage human reading time to extract keyphrases from microblog posts. Experiments show that the proposed models yield better performance on two microblog datasets. |
| | *Ikhwantri et al., 2023* | *Multi-task* | This paper provides the first broad overview of the relation between different interpretation methods and human eye-movement behavior across different tasks. |
| Study of EEG signals and Eye-tracking signals based NLP | *Zheng et al., 2014* | *Emotion recognition* | The best average accuracies based on EEG signals and eye tracking data are 71.77% and 58.90%, respectively. |
| | *Puma et al., 2014* | *Multi-tasks* | This work proved that it's feasible to partition EEG data based on eye-tracking data so as to increase to signal to noise ratio. |
| | *Hollenstein et al., 2019b* | *Multi-tasks* | The gaze and EEG signals of humans reading text, show great potential in improving NLP tasks and facilitate insights into language processing. |
| | *Gupta et al., 2020* | *visual perception* | The compiled data confirms distinct eye movement characteristics for real vs fake videos, and utility of the eye-track saliency maps for spatial forgery detection. |
| | *Daza et al., 2021* | *attention level estimation* | The proposed mEBAL improves previous databases in terms of acquisition sensors and samples. |

**Table 3.** Summary of representative studies of EEG and eye-tracking signals in NLP

EEG signals also play an important role in language perception tasks. Sarrett et al.(2020) elucidated the brain's processing mechanisms of fine-grained speech perception using EEG signals, which provided more details and facts for the study of cortical language networks. Lum et al.(2022) investigated the relationship between spontaneous neural oscillatory activity and children's language skills during a sentence repetition task. The findings revealed a negative correlation between EEG signals in the theta frequency band and task performance. Besides, EEG signals can assist in the analysis of semantics and syntax of texts through deep learning models(Keles et al., 2023). Serving as a valuable tool for gaining deeper insights into human cognitive processing while reading, EEG has been applied in various research domains, including language and motion perception (Scaltritti et al., 2020), cognitive ability assessment (De Vito et al., 2018), cross-linguistic studies (León Rodríguez et al., 2022) and comparative investigations of language and non-language cognitive processing mechanisms (Dudschig, 2022).

**Study of Eye-tracking signals based NLP.** Human reading behavior, to some extent, reflects cognitive processes. Eye-tracking behavior, as a common behavior in reading, which has been the subject of research since the late 19th and early 20th centuries. Eye-tracking technology primarily examines eye movement patterns during both graphic scanning and text reading. However, due to its primitive nature and substantial errors, it did not gain popularity as a mainstream research method. As eye-tracking technology has gradually matured, an increasing number of studies in NLP have focused on analyzing eye-tracking data. Presently, this technique finds widespread application in research related to keyphrase extraction (Zhang & Zhang, 2021), emotion recognition (Geraets et al., 2021), online comment analysis (Jin et al., 2023), and other upstream and downstream investigations.

**Study of EEG signals and Eye-tracking signals based NLP.** The early research on the integration of eye-tracking signals and EEG signals primarily centered around sleep studies and the treatment of mental disorders like epilepsy and depression (Li et al., 2015). Subsequently, its application expanded to encompass human cognition research. Advancements in neuroscience theories and technologies have led to the development of sophisticated signal processing techniques. Scholars have identified artifact removal as a critical step in integrating two types of signals. Plöchl et al. (2012) confirmed the diverse origins of ocular artifacts based on electrode placement, gaze direction and reference selection. They also proposed a targeted identification and correction technique for ocular artifacts. The integration of eye-tracking signals and EEG signals for emotion recognition is a recent research hotspot (Zheng et al., 2014). Some scholars have worked on constructing comprehensive cognitive processing corpora that incorporate both eye-tracking signals and EEG signals to facilitate research across different disciplines. Hollenstein et al.(2018) developed the cognitive language processing corpus, ZUCO, serving as a primary data source for this study. Moreover, several open-source cognitive corpora encompassing multiple cognitive signals have emerged. For instance, the mEBAL database (Daza et al., 2021) contains EEG frequency bands, blink events, facial expressions and gesture events to analyze user attention levels. Another corpus, the FakeET database (Gupta et al., 2020), is built for perceptual depth-faking videos.

In summary, current research on AKE from Microblogs has mainly concentrated on approaches relying on internal textual features. Nevertheless, due to the unstructured property of Microblog posts, achieving substantial improvements in model performance is difficult. Consequently, researchers have increasingly incorporated external features generated during human reading, like cognitive signals, into Microblog-based AKE studies. Among these cognitive signals, eye-tracking signals are widely used, as they offer essential insights into human attention allocation. It is noted that Zhang & Zhang (2021) demonstrated the beneficial impact of eye-tracking signals on AKE from Microblogs by considering human readings time. In contrast, we introduces a groundbreaking approach by integrating EEG signals, which directly reflect human thinking activities, with eye-tracking signals for AKE from Microblogs. The incorporation of EEG signals not only provides minimal latency but also enhances interpretability. Thus, we are able to delve into more intricate cognitive processes like logical reasoning, relationship recognition, and emotional fluctuations. Through various frequency bands of EEG signals, we gain profound insights into human cognitive activities during text reading, tapping into the association between EEG rhythms and brain functional networks. This integrated approach shows immense promise in advancing AKE from Microblogs, as it enables a deeper understanding of the cognitive mechanisms underlying information extraction from vast and sparsely-structured Microblogs.

## 3. Methodology

We organized the research framework of this study according to the research process (Refer to Fig. 1). It consists of four specific steps: The first step is to construct the "Multi-Source Cognitive Signals-based AKE Corpus". This corpus is synthesized by correlating word-level EEG and ET signals from the Zuco dataset with AKE content from the General-Twitter and Election Trec datasets. The second step entails building the AKE models, including baseline models, attention-based models, PLM and LLM-Based models. In the third step, we consider how to effectively combine cognitive signals with AKE models, and different combination strategies are formulated based on the structures of different AKE models. Finally, we conducted AKE experiments with cognitive signals, comprehensively investigating the impact of cognitive signals on the performance of AKE tasks through tests in both whole-band and sub-frequency domains. Section 3.1 presents the dataset and specific features of cognitive signals, while Section 3.2 elaborates on the components of the AKE from Microblogs framework.

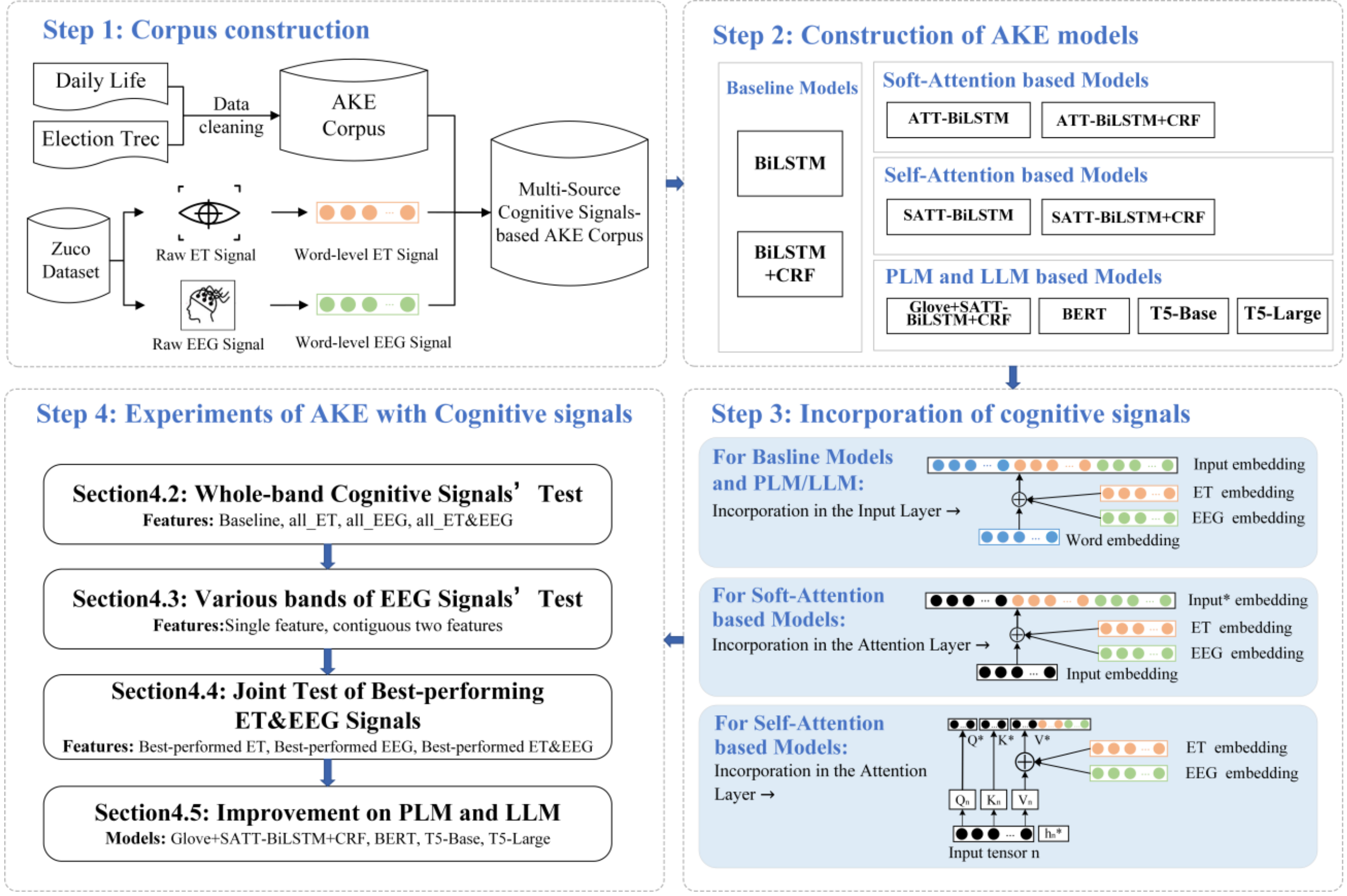

**Fig. 1. Research Framework Diagram for AKE from Microblogs**

### 3.1. Dataset

In our study, two kinds of data are used: the AKE from Microblogs data and the cognitive signal data from human readings behaviors. The cognitive signal data includes data from two sources: eye tracking signals and EEG signals. Training and testing data were obtained by matching and concatenating word-level eye-tracking and electroencephalogram (EEG) features from the cognitive signal dataset with word vectors from the AKE from Microblogs dataset. Unmatched words were considered out-of-vocabulary (OOV) words, and their feature values were set to the average value of that feature across the entire ZUCO corpus.

#### 3.1.1. Cognitive Signals Dataset

Cognitive signals, as a widely utilized data source in recent NLP research, facilitate a more profound understanding of the cognitive and neural mechanisms that underlie human readings behavior. In this study, we choose the Zurich Cognitive Language Processing Corpus (ZUCO)[1], which captures eye-tracking signals and EEG signals of 12 adult native speakers reading approximately 1100 English sentences in normal and task reading modes. Only data from the normal reading mode were utilized to align with human natural reading habits. The reading corpus includes two datasets: 400 movie reviews from the Stanford Sentiment Treebank and 300 paragraphs about celebrities from the Wikipedia Relation Extraction Corpus. After preprocessing, 17 Eye-tracking features and 8 EEG features were extracted from the dataset.

**(1) Eye-tracking signals**

| Type of Features | Eye-tracking Features(abbreviations) | Explanation of Features |
|---|---|---|
| Early-Stage Features | First Fixation Duration (FFD) | The duration of word W that is first fixated |
| | First Pass Duration (FPD) | The sum of the fixations before eyes leave the word W |
| Late-Stage Features | Number of Fixations (NFIX) | The number of times word W that is fixated |
| | Fixation Probability (FP) | The probability that word W is fixated |
| | Mean Fixation Duration (MFD) | The average fixation duration for word W |
| | Total Fixation Duration (TFD) | The total duration of word W that is fixated |
| | N Re-Fixations (NR) | The number of times word W that is fixated after the first fixation |
| | Re-Read Probability (RRP) | The probability of word W that is fixated more than once |
| Contextual Features | Total Regression-from Duration (TRD) | The total duration of regressions from word W |
| | W-2 Fixation Probability (W-2 FP) | The fixation probability of the word W-2 |
| | W-1 Fixation Probability (W-1 FP) | The fixation probability of the word W-1 |
| | W+1 Fixation Probability (W+1 FP) | The fixation probability of the word W+1 |
| | W+2 Fixation Probability (W+2 FP) | The fixation probability of the word W+2 |
| | W-2 Fixation Duration (W-2 FD) | The fixation duration of the word W-2 |
| | W-1 Fixation Duration (W-1 FD) | The fixation duration of the word W-1 |
| | W+1 Fixation Duration (W+1 FD) | The fixation duration of the word W+1 |
| | W+2 Fixation Duration (W+2 FD) | The fixation duration of the word W+2 |

**Table 4. Summary of Eye-Tracking Features**

Eye-tracking signals record subjects' eye movements, such as fixation, saccade, and regressions, during the reading process. In ZUCO Corpus, Hollenstein et al.(2019) categorized the 17 eye-tracking features into three groups: *Early-Stage Features*, *Late-Stage Features* and *Contextual Features*, encompassing all gaze behavior stages and contextual influences. Early-Stage Features reflect readers' initial comprehension and cognitive processing of the text, while Late-Stage Features indicate readers' syntactic and semantic comprehension. Contextual Features refer to the gaze behavior of readers on the words surrounding the current word, as demonstrated in Table 4.

**(2) EEG signals**

EEG is a bio-electrical signal measurement used to assess brain activity by detecting electrical potential changes in brain neurons through multiple scalp electrodes. **EEG spectral analysis**, or EEG rhythm analysis, is a widely utilized EEG analysis method in various scientific disciplines. It was confirmed by Hans Berger that different frequencies recorded in the EEG reflect different states of the brain. Through extensive experimentation, it is now widely accepted that EEG signals can be divided into different frequency bands (or rhythms), each with distinct functions and characteristics.

EEG signals can be typically classified into θ, α, β and γ rhythms in the frequency domain (Kim & Im, 2018). These rhythms exhibit distinct activity patterns during human readings. Studies have linked increased θ rhythm (4-8Hz) during human readings to language processing and memory encoding. α rhythm (8-13Hz) is associated with attention regulation (Van Diepen et al., 2019) and reduced alpha rhythm has been associated with improved reading efficiency (Proverbio et al., 2004). β rhythm (13-30Hz) is linked to higher-order cognitive processing (X. Wang et al., 2022), while low-frequency β band signal is crucial for human executive control (Lundqvist et al., 2023). γ rhythm (30-100Hz) is associated with perception and attention (Demirel et al., 2012). Furthermore, it has been revealed that mutual interactions between β and γ rhythms, with distinct roles in cognitive processing stages and mutual regulation contributing to advanced cognitive activities in human's brain.

For our study, the recorded EEG signals used a 128-channel neural signal acquisition system, categorized into four frequency bands with two features per band (refer to Table 5 for details). EEG signals were synchronized with "shared" cognitive time points using the "EYE-EEG" toolbox[2], with all synchronization errors within 2 milliseconds (i.e., one data point). This tight synchronization is crucial for accurate alignment of EEG signals with eye-tracking data, allowing us to precisely examine the cognitive activities during text reading with minimal temporal discrepancies.

| EEG Rhythm(Features) | Frequency（Hz） |
|---|---|
| $\theta_1$ (EEG1) | 4~6 |
| $\theta_2$ (EEG2) | 6.5~8 |
| $\alpha_1$ (EEG3) | 8.5~10 |
| $\alpha_2$ (EEG4) | 10.5~13 |
| $\beta_1$ (EEG5) | 13.5~18 |
| $\beta_2$ (EEG6) | 18.5~30 |
| $\gamma_1$(EEG7) | 30.5~40 |
| $\gamma_2$(EEG8) | 40~49.5 |

**Table 5. Summary of EEG Features**

#### 3.1.2. Microblog-based AKE Dataset

The data of Microblog-based AKE includes two datasets: the General-Twitter dataset and the Election-Trec dataset. The General-Twitter dataset[3], developed by (Zhang et al., 2016), employs Hashtags as keyphrases for each tweet. It consists of 78,760 training tweets and 33,755 testing tweets, with an average sentence length of about 13 words. The Election-Trec dataset[4] is derived from the open-source dataset TREC2011 track4. After removing all "#" symbols, it contains 24,210 training tweets and 6,054 testing tweets (refer to Table 6 for details). As shown in Fig. 7, we computed the statistical information for both cognitive data and AKE data. Specifically, it includes the total number of sentences, the total number of words, the vocabulary size (i.e., the total count of unique words), and the Coverage rate of Zuco (i.e., the percentage of overlap between the Zuco vocabulary and the vocabulary of the AKE dataset).

| Data Type | Dataset | Sentence(s) | Word(s) | vocab | Coverage rate of Zuco（%） |
|---|---|---|---|---|---|
| Cognitive Data | ZUCO | 739 | 15138 | 4050 | |
| Twitter Data | Election-Trec | 30264 | 602785 | 33583 | 8.579 |
| | General-Twitter | 112515 | 1775694 | 82736 | 4.038 |

**Table 6. Statistical information of Cognitive Data and AKE Data**

### 3.1.3. Multi-Source Cognitive Signals-based AKE Corpus

As presented in Table 7, the *Multi-Source Cognitive Signals-based AKE Corpus* is a comprehensive corpus that combines word-level cognitive signals (from Zuco dataset) and AKE corpora (from General-Twitter and Election-Trec datasets). The reason for integrating these two types of data is that the cognitive information provided by ET and EEG signals may has the potential to enhance the context comprehension ability of the AKE model, thus better revealing the recognition and processing of keyphrases during the reading process.

| Words | ET Signals | | | | | EEG Signals | | | | | Tag |
|---|---|---|---|---|---|---|---|---|---|---|---|
| | ET1 | ET2 | …… | ET16 | ET17 | EEG1 | EEG2 | …… | EEG7 | EEG8 | |
| i | 0.685216 | 2.242174 | | 2.284111 | 2.315416 | 2.904706 | 0.000263 | | 3.452123 | 3.173984 | O |
| nominate | 0.685216 | 2.242174 | | 2.284111 | 2.315416 | 2.904706 | 0.000263 | | 3.452123 | 3.173984 | O |
| NAME | 0.685216 | 2.242174 | | 2.284111 | 2.315416 | 2.904706 | 0.000263 | | 3.452123 | 3.173984 | O |
| NAME | 0.685216 | 2.242174 | | 2.284111 | 2.315416 | 2.904706 | 0.000263 | | 3.452123 | 3.173984 | O |
| ariana | 0.685216 | 2.242174 | | 2.284111 | 2.315416 | 2.904706 | 0.000263 | | 3.452123 | 3.173984 | B |
| rilakkuma | 0.685216 | 2.242174 | | 2.284111 | 2.315416 | 2.904706 | 0.000263 | | 3.452123 | 3.173984 | I |
| contest | 1 | 3 | | 0 | 3 | 5 | 4 | | 5 | 5 | E |
| 0 | 0.222222 | 1.888889 | | 2.611111 | 2.222222 | 2.333333 | 1.944444 | | 2.833333 | 2.555556 | O |
| | | | | | | | | | | | |
| delay | 0.685216 | 2.242174 | | 2.284111 | 2.315416 | 2.904706 | 0.000263 | | 3.452123 | 3.173984 | O |
| alert | 0.685216 | 2.242174 | | 2.284111 | 2.315416 | 2.904706 | 0.000263 | | 3.452123 | 3.173984 | O |
| - | 0 | 0.071429 | | 1.964286 | 1.607143 | 0.107143 | 0.071429 | | 0.142857 | 0.178571 | O |
| rouge | 0.685216 | 2.242174 | | 2.284111 | 2.315416 | 2.904706 | 0.000263 | | 3.452123 | 3.173984 | O |
| 0 | 1.259067 | 3.046632 | | 1.989637 | 2.373057 | 3.259067 | 2.715026 | | 3.906736 | 3.626943 | O |

Table 7. **Data sample of *Multi-Source Cognitive Signals-based AKE Corpus***

**Note**: Blank lines are used to split sentences. All of the cognitive values are normalized to the 0~9.

Initially, a vocabulary $V_Z$ is formed based on the Zuco dataset. Subsequently, a list of cognitive features is built for each word, and the cognitive signal value for each word is determined as the average feature value of its occurrences in $V_Z$. For OOV words, we calculated their value by summing their respective values under a specific signal and taking the average. The complete cognitive vocabulary $V_C$ was formed by concatenating the cognitive feature list between each word and its corresponding tag. Lastly, the words in the AKE corpora were matched with the vocabulary $V_C$. If a word does not appear in $V_C$, the feature list for OOV words would be concatenated after the word.

### 3.2. AKE model from Microblogs with EEG signals and eye-tracking signals

In this study, the AKE task is treated as a sequence labeling task. To consider semantic information in the text context, we choose BiLSTM as baseline models. Additionally, we integrate two improved attention mechanisms to assess the influence of cognitive features on AKE from Microblogs. The attention layer combines eye-tracking and EEG signal features as external features, which are concatenated with the input sequence of the Soft Attention Mechanism(Bahdanau et al., 2016) and the query vector of the Self-Attention Mechanism(Vaswani et al., 2017), directly incorporating cognitive feature information into the model.

### 3.2.1. Baseline models

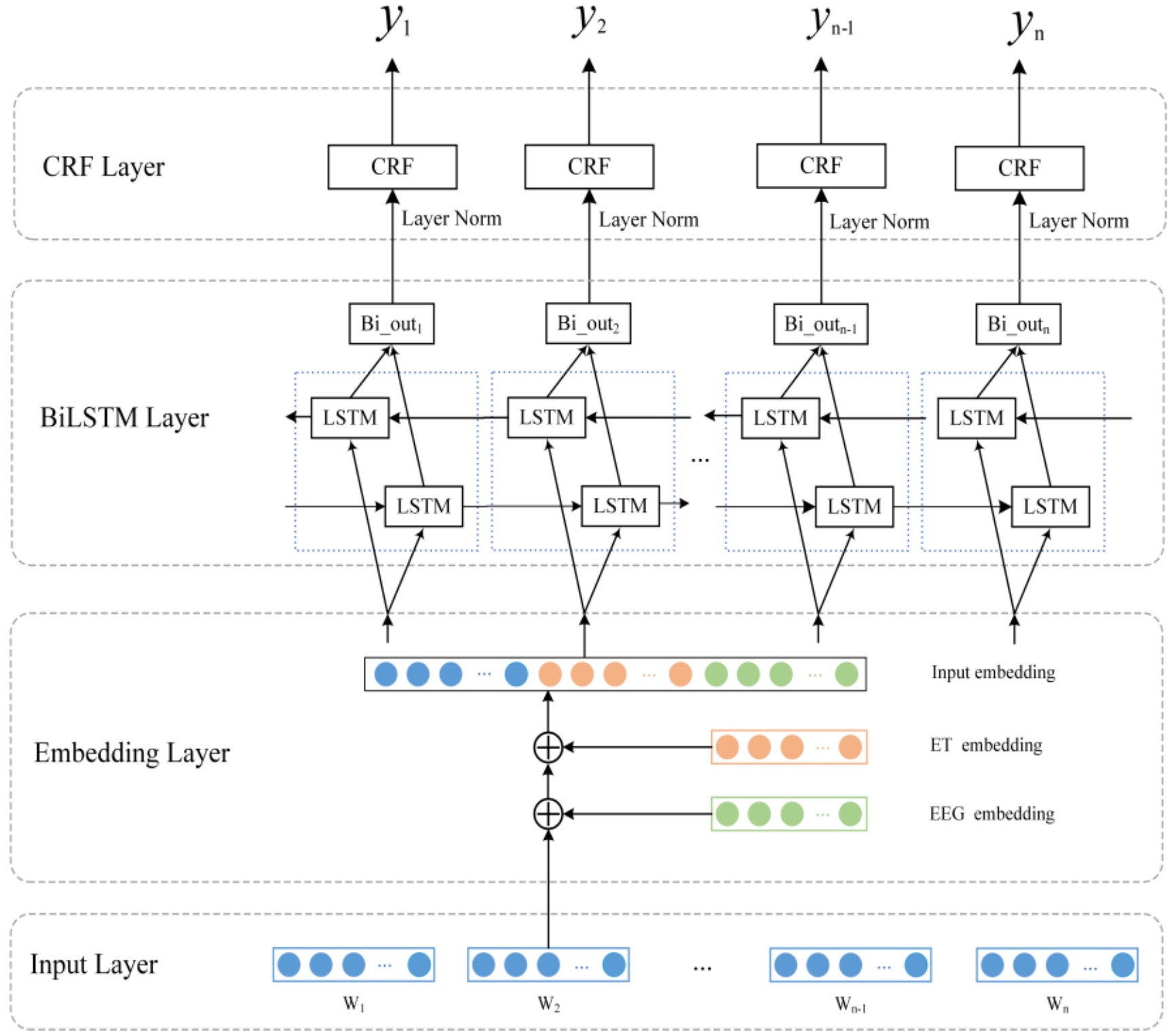

**Fig. 2. Structure of BiLSTM+CRF with ET and EEG features**

We selected The BiLSTM model and BiLSTM+CRF model as baseline models for two primary reasons. On the one hand, Zhang & Zhang (2021) used the BiLSTM model as a baseline for their AKE task based on human reading time. Therefore, we initially trained our reading cognitive signals on BiLSTM model to obtain a direct comparison of AKE performance improvements between two studies. On the other hand, in subsequent experiments, we separately integrated two distinct attention mechanisms into the BiLSTM+CRF model. To distinguish AKE models with attention mechanisms, we included the BiLSTM+CRF model in our baseline models.

Fig. 2 illustrates the structure of the BiLSTM+CRF model with the integration of eye-tracking signals and EEG signals. In the baseline model, for a given target tweet $x_i$ represented by a word sequence $< w_1, w_2, \dots, w_n >$ with a total length of $n$, we concatenate word-level eye-tracking and EEG features with word embeddings at the Embedding layer to obtain the input vector $< vec(w_i), vec(ET_{i,1}), \dots, vec(ET_{i,17}), vec(EEG_{i,1}), \dots, vec(EEG_{i,8}) >$ for the BiLSTM layer.

The feature vector sequence $V = < v_1, v_2, \dots, v_n >$ is fed into the BiLSTM layer to generate predicted probabilities for each class label. Subsequently, the LayerNorm normalization function is applied to ensure that the model's gradients are within an appropriate range, stabilizing the hidden states. The output feature matrix undergoes linear transformation before being passed to the CRF layer for calculating the model's loss value and the probabilities associated with each label $Y_i = y_1, y_2, \ldots, y_{|x_i|}$.

### 3.2.2. Attention-based models

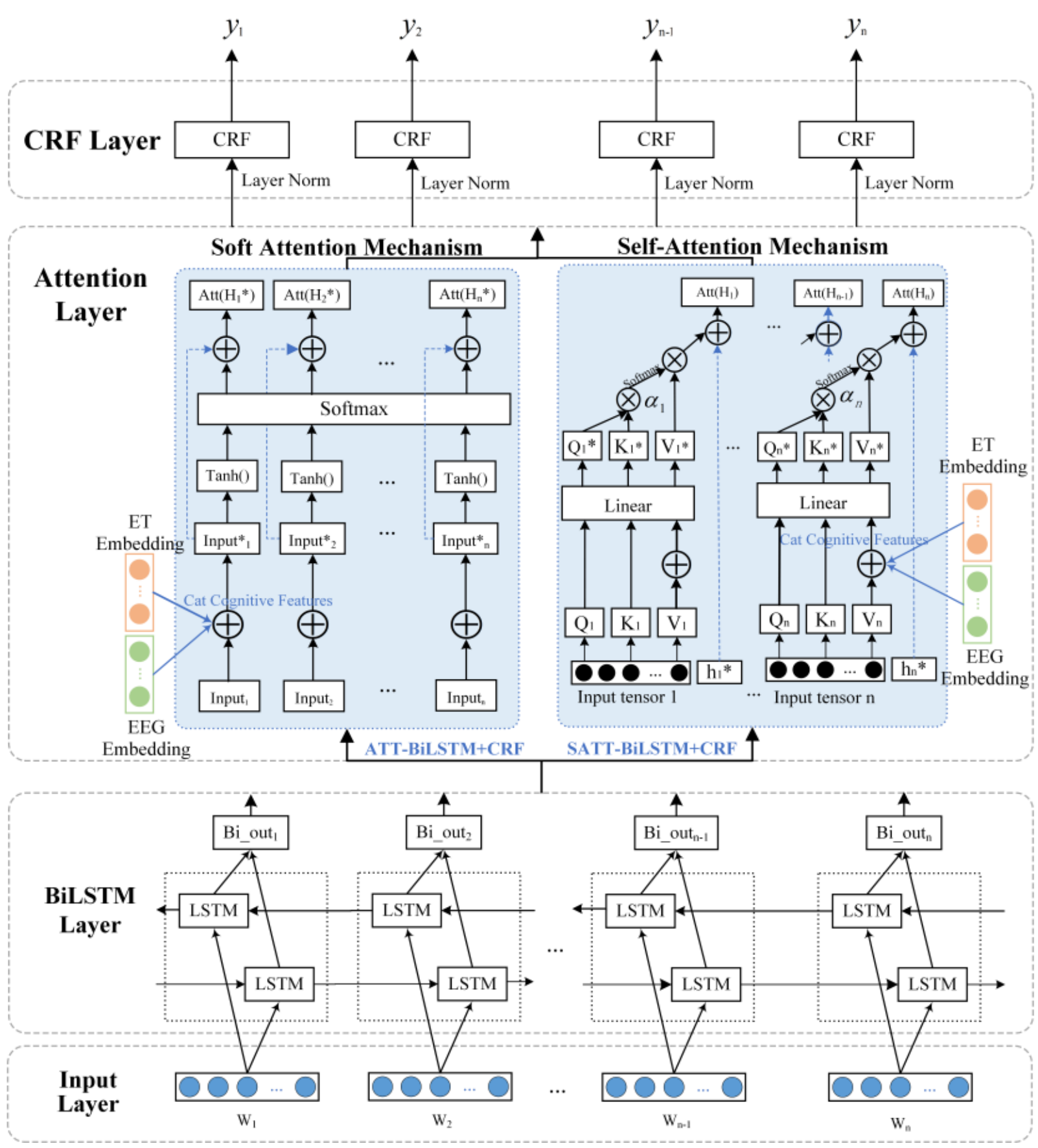

**Fig. 3. Structure of BiLSTM+CRF combining attention mechanisms with ET and EEG features**

The attention mechanism is a significant tool in NLP, enabling models to focus on important parts of the sequence data and improve predictions. In this study, we integrate two attention mechanisms: the soft attention layer and the self-attention layer, represented by light blue boxes in the figure below (Refer to Fig. 3 For details). After the BiLSTM layer, the results feed into the left blue box for the ATT-BiLSTM+CRF model or the right blue box for the SATT-BiLSTM+CRF model. To preserve cognitive signals' significance, we concatenate the embeddings of eye-tracking signals and EEG signals with the original word embeddings at the input of the soft attention layer and the query vector of the self-attention layer.

**(1) Soft attention mechanisms**

Soft attention mechanisms are particularly useful in scenarios where models need to assign varying attention weights to different positions within a sequence. In our task, concatenating cognitive features with the input $H = < h_1, h_2 \ldots, h_n >$ at the attention layer allows the model to directly learn the features conveyed by eye-tracking and EEG signals. The soft attention mechanism entails weighted multiplication of the input matrix with the feature matrix obtained through activation functions' normalization. Equation (1) represents the attention score function, with $\omega^T$ denoting the transpose of the parameter vector learned during training. After normalization using Equation (2), the attention allocation value α for a specific time step is obtained. The final output of the soft attention mechanism layer is obtained through the weighted averaging of all attention scores at each time step, as showed in Equation (3). Soft attention mechanism provides flexibility in capturing the importance of different cognitive features, leading to improved performance in AKE from Microblogs.

$$M = \tanh(H^*) \tag{1}$$

$$\alpha = softmax(\omega^T M) \tag{2}$$

$$att(H^*) = \sum_{i=1}^{N} H^*_i \, \alpha_i \tag{3}$$

**(2) Self-attention mechanisms**

The self-attention mechanism, as a special type of soft attention mechanism, is particularly well-suited for tasks like language modeling and text classification. It treats each input position as a key-value pair and a query vector to compute correlations with all other positions. K, Q and V represent the input sequence length, batch size, and word vector length, respectively.

$$\mathrm{K} = W^K X \tag{4}$$

$$Q = W^Q X \tag{5}$$

$$\mathrm{V} = W^V X \tag{6}$$

In contrast to the baseline model, we append the cognitive signals after the query vector $< \mathrm{V}, \mathrm{vec}(ET_1), \ldots, \mathrm{vec}(ET_{17}), \mathrm{vec}(EEG_1), \ldots, \mathrm{vec}(EEG_8) >$ resulting in a more substantial improvement compared to concatenating at the word embedding layer. We first compute the attention scores for each word, and subsequently obtain the weighted vector representation of all positions by weighting each position's vector.

$$Attention(Q_i, K_j, V_j) = soft\max(\frac{Q_i K_j^T}{\sqrt{d_k}})V_j \quad (7)$$

$$output_i = \sum_{j=1}^{n} attention(K_i, Q_j, V_j) \quad (8)$$

### 3.2.3. Improvement models based on PLM and LLM

Pre-trained language models (PLMs) adopt the "pre-training and fine-tuning" learning paradigm, wherein context-aware word representations are obtained by pre-training on large-scale unlabeled corpora with specifically designed pre-training tasks. These representations have notably enhanced the performance of NLP tasks and have spurred significant research activity in this domain. With the advancement and widespread adoption of ChatGPT, research in NLP pertaining to Large Language Models (LLMs, i.e. PLM which contains tens or hundreds of billions of parameters) has seen a rapid proliferation. In this section, we propose three improvement strategies to enhance the performance of AKE from Microblogs. The first strategy combines Glove Embeddings with the SATT-BiLSTM+CRF model. The second strategy involves the use of the BERT model, while the final one utilizes the T5 (T5-Base, T5-Large) model for AKE from Microblogs.

**(1) Glove-based SATT-BiLSTM+CRF**

The GLOVE model, introduced in 2014 by Stanford Professor Manning et al., is a word vector training model obtained through extensive pre-training on large-scale corpora. We incorporate Glove's word vectors into the embedding layer of the SATT-BiLSTM+CRF model to enhance our ability to capture the semantic relationships among words, thereby refining the model's performance.

**(2) BERT**

Bert is a pre-trained language model based on the encoder architecture in Transformer. In our study, we consider AKE as a sequence labeling task. During the fine-tuning of Bert, every word at each position in the original input sequence is transformed into a 768-dimensional vector after passing through the Bert model. Consequently, a straightforward linear classifier is attached to the last layer of Bert, converting the 768 dimensions into predictions corresponding to the number of labels, thus yielding the forecasted results.

**(3) T5**

T5, an acronym for "Text-to-Text Transfer Transformer," represents a LLM grounded in the Transformer architecture. T5-Base constitutes the standard model in T5 series, endowed with relatively fewer parameters. In contrast, T5-Large boasts a more extensive parameter set, typically necessitating more substantial computational resources for training and deployment. The T5 model undergoes pre-training on vast text corpora, endowing it with formidable linguistic comprehension capabilities, a quality particularly advantageous when handling AKE corpora combined with cognitive signals. In parallel to BERT's methodology, we conducted AKE tests separately using T5-Base and T5-Large.

## 4. Experiments and Results

This section describes the experimental parameter settings and the specific experiments and results of AKE from Microblogs. Section 4.1 presents the description of experimental parameters and evaluation methods. For the experiments, in Sections 4.2 to 4.5, we conduct various tests to evaluate the performance of our AKE model with different feature combinations and attention mechanisms. Section 4.2 is a test of the overall performance of AKE with cognitive signals, Section 4.3 focuses on AKE tests using fine-grained EEG signals, and Section 4.4 integrates eye-tracking signals and EEG signals for a joint AKE test. In Section 4.5, we further propose an improved scheme based on the pre-trained model. Finally, we provide an example study on AKE testing in Section 4.6.

### 4.1. Experimental setup

We implemented the AKE framework using PyTorch with default experimental parameters. The word embedding dimension was set to 128, except for the improved model using pre-trained GloVe word vectors with 100 dimensions. The training model was based on a 256-dimensional BiLSTM network as the baseline model. All models underwent five training iterations (n=5), each comprising 20 epochs to mitigate the impact of random fluctuations. Hyper parameters were fine-tuned to optimize performance on each dataset and each model. Evaluation involved calculating the best $F_1$ score and the corresponding P (precision) and R (recall) values for the top 20 epochs and averaging the results across the five iterations (as showed in Equations 9-11). $Y_{pred}$ represents the predicted set of keywords, while $Y_{gold}$ represents the ground truth set.

$$\mathrm{P@k} = \frac{1}{n} * \frac{Y_{pred} \cap Y_{gold}}{|Y_{pred}|} \tag{9}$$

$$\mathrm{R@k} = \frac{1}{n} * \frac{Y_{pred} \cap Y_{gold}}{|Y_{gold}|} \tag{10}$$

$$F_1@k = \frac{2}{n} * \frac{P@k * R@k}{P@k + R@k} \tag{11}$$

### 4.2. Overall Performance Test of AKE with human reading cognitive signals

| Features | Models | Election-Trec | | | General-twitter | | |
|---|---|---|---|---|---|---|---|
| | | P | R | $F_1$ | P | R | $F_1$ |
| - | Baseline* | 69.7 | 58.71 | 61.67 | 79.33 | 69.16 | 70.67 |
| | BiLSTM+CRF | 73.99 | 64.95 | 64.64 | 86.09 | 80.95 | 81.63 |
| | ATT-BiLSTM | 68.72 | 56.82 | 62.13 | 83.56 | 76.65 | 78.64 |
| | ATT-BiLSTM+CRF | 72.99 | 63.87 | 63.25 | 87.59 | 79.08 | 79.93 |
| | SATT-BiLSTM | 74.67 | 60.52 | 61.59 | 83.56 | 68.06 | 77.49 |
| | SATT-BiLSTM+CRF | **81.84** | **66.83** | 65.38 | 86.18 | 80.85 | 81.74 |
| All_EEG | Baseline* | 67.43 | 57.61 | 62.53 | 81.83 | 72.19 | 74.74 |
| | BiLSTM+CRF | 74.41 | 66.18 | 65.9 | **87.78** | 81.37 | 81.79 |
| | ATT-BiLSTM | 69.18 | 59.1 | 63.73 | 84.56 | 77.69 | 79.26 |
| | ATT-BiLSTM+CRF | 72.42 | 63.84 | 64.09 | 84.34 | 79.17 | 80.64 |
| | SATT-BiLSTM | 76.48 | 61.81 | 63.22 | 84.12 | 76.64 | 78.51 |
| | SATT-BiLSTM+CRF | 72.6 | 65.87 | **67.82** | 84.72 | 81.61 | **82.46** |
| All_ET | Baseline* | 67.43 | 57.66 | 62.16 | 85.7 | 68.83 | 74.52 |
| | BiLSTM+CRF | 74.35 | 65.04 | 65.31 | 86.54 | 81.11 | 81.64 |
| | ATT-BiLSTM | 65.67 | 60.27 | 62.83 | 84.55 | 77.92 | 78.66 |
| | ATT-BiLSTM+CRF | 72.36 | 63.95 | 63.92 | 84.91 | 79.55 | 80.47 |
| | SATT-BiLSTM | 79.84 | 59.37 | 62.55 | 86.38 | 76.16 | 78.98 |
| | SATT-BiLSTM+CRF | 72.71 | 65.78 | 67.52 | 84.53 | 81.61 | 82.22 |
| All_EEG & All_ET | Baseline* | 68.2 | 56.77 | 61.96 | 85.39 | 69.62 | 74.15 |
| | BiLSTM+CRF | 73.96 | 65.18 | 65.18 | 86.33 | 81.05 | 81.34 |
| | ATT-BiLSTM | 68.92 | 58.59 | 63.32 | 84.90 | 77.54 | 79.05 |
| | ATT-BiLSTM+CRF | 72.54 | 63.25 | 64 | 84.47 | 78.81 | 80.33 |
| | SATT-BiLSTM | 81.24 | 62.15 | 63.15 | 84.85 | 76.59 | 78.18 |
| | SATT-BiLSTM+CRF | 72.03 | 65.57 | 67.66 | 84.6 | **81.73** | 82.26 |

**Table 4. $F_1$ Value of AKE augmented by all Eye-Tracking Signals(All_ET), all EEG Signals(All_EEG) and both(All_ET& All_EEG)**

**Note**: Bold and underlined annotations indicate the best scores for each column and each feature combination, respectively. In the "Models" column, "Baseline*" denotes the baseline model Bi-LSTM.

In the first test, we compared the performance of the AKE under four feature combinations: "-," "All_EEG," "All_ET," and "All_EEG& All_ET" (Refer to Table 4 for details). "-" indicates the model without using any cognitive processing signals. "All_EEG" and "All_ET" represent the model with only EEG signals and only eye-tracking signals, respectively. "All_EEG&All_ET" indicates the model that combines both EEG and eye-tracking signals simultaneously. Upon analyzing the results, we observed that:

Firstly, both the attention mechanism and the CRF layer improved the performance of the AKE model from Microblogs. By concatenating the cognitive signals with the input matrix or its query vector in the attention layer, the model learns the information from textual information and cognitive signals better. Comparing the results from the table, the inclusion of attention mechanisms and CRF further both enhances the performance of the model. The best performance is attained when combining both self-attention and CRF layers, signifying substantial improvements in the model's learning capacity with these two structures.

Secondly, the performance of both the baseline model and the model with CRF and attention mechanism is further improved by incorporating cognitive signals, indicating a stable enhancement effect on AKE. Among different combinations of features, the model incorporating EEG signals achieves optimal performance. The enhancement effect of EEG signals is superior to that of eye-tracking signals, while the combined result is intermediate between the individual results. This might be attributed to the low signal-to-noise ratio of EEG signals, which results in slightly lower performance when both signals are combined than when EEG signals are used alone.

Among six models in AKE framework, SATT-BiLSTM+CRF outperformed others, followed by BiLSTM+CRF. Notably, SATT-BiLSTM+CRF showed substantial improvements in $F_1$ scores, with 3.71 and 11.07 increases on the Election-Trec dataset and the General-Twitter dataset, respectively. These results highlight the significant enhancement effect of the self-attention mechanism and CRF, particularly on the larger-scale General-Twitter dataset.

Finally, in the absence of the CRF layer, the soft attention mechanism shows more pronounced improvements in AKE from Microblogs compared to the self-attention mechanism. However, when the CRF layer is included, the results are reversed. We speculate that some structures of the soft attention mechanism and the CRF layer might have the same function, leading to a decrease in model performance when both are combined.

### 4.3. AKE Performance Test of EEG signals in different frequency bands

EEG's excellent temporal resolution allows capturing neuronal activity changes in the order of tens of milliseconds. The aforementioned studies on cognitive function-related EEG rhythms will be invaluable evidence for analyzing the performance of EEG frequency bands in this paper. Table 5 shows a comparative test of AKE from Microblogs using individual EEG signals and assessed the performance differences by combining adjacent pairs of EEG signals. The feature combinations can be categorized into two groups: individual EEG features and adjacent pairs of EEG features.

| Feature Combination | Election-Trec | | | | General-Twitter | | | |
|---|---|---|---|---|---|---|---|---|
| | Baseline* | ATT-BiLSTM | BiLSTM +CRF | SATT-BiLSTM +CRF | Baseline* | ATT-BiLSTM | BiLSTM +CRF | SATT-BiLSTM+ CRF |
| - | 61.61 | 62.13 | 64.64 | 65.38 | 70.67 | 78.64 | 81.63 | 81.74 |
| All_EEG | 62.53 | **63.73** | **65.9** | **67.82** | 74.74 | 79.26 | 81.79 | 82.46 |
| EEG1 | 61.91 | 62.35 | 64.54 | 67.53 | 74.64 | 79.01 | 81.73 | 82.3 |
| EEG2 | 61.9 | 62.26 | 65.24 | 67.47 | 72.59 | 78.7 | 81.73 | 82.2 |
| EEG3 | 61.83 | 62.09 | 64.91 | 67.32 | 72.58 | 78.6 | 81.74 | 82.3 |
| EEG4 | 62.06 | 62.05 | 64.9 | 67.61 | 74.41 | 79.01 | 81.78 | 82.28 |
| EEG5 | **62.53** | 62.51 | 65.71 | 67.83 | **74.84** | 79.09 | **81.89** | 82.41 |
| EEG6 | 62.15 | 62.23 | 65.44 | 67.66 | 74.53 | 78.91 | 81.72 | 82.36 |
| EEG7 | 62.21 | 62.31 | 64.93 | 67.7 | 74.72 | 78.94 | 81.80 | 82.27 |
| EEG8 | 62.24 | 62.25 | 65.15 | 67.49 | 74.8 | 79.02 | 81.83 | 82.3 |
| EEG1,2 | 62.11 | 62.37 | 65.34 | 65.34 | 74.03 | 79.03 | 81.76 | 82.21 |
| EEG2,3 | 62.23 | 62.1 | 65.22 | 65.21 | 72.57 | 78.84 | 81.71 | 82.27 |
| EEG3,4 | 62.21 | 62.23 | 65.5 | 65.23 | 73.73 | 79.16 | 81.73 | 82.29 |
| EEG4,5 | 62.5 | 62.14 | 65.08 | 65.47 | 74.44 | 79.13 | 81.74 | 82.26 |
| EEG5,6 | 62.57 | 62.59 | 65.62 | 65.79 | 74.71 | **79.27** | 81.88 | **82.49** |
| EEG6,7 | 62.32 | 62.42 | 64.84 | 65.05 | 74.48 | 79.17 | 81.64 | 82.41 |
| EEG7,8 | 62.13 | 62.1 | 65.29 | 65.35 | 73.83 | 79.22 | 81.8 | 82.3 |

**Table 5. $F_1$ Value of AKE model augmented by Single EEG Signal and Contiguous EEG Signals**

As depicted in Fig. 4, we compared the $\Delta F_1$ of different EEG rhythm combinations in the Election-Trec and General-Twitter datasets. Distinct colors indicate different EEG rhythms: green for θ rhythm, blue for α rhythm, yellow for β rhythm, and orange for γ rhythm. The gradient colors represent combinations that encompass both end colors of the respective rhythms. The model performance is significantly enhanced when combining individual EEG features, with the low-frequency β band (EEG5) showing the most prominent improvement, followed by high-frequency β band, low-frequency γ band, and high-frequency gamma band. When combining contiguous pairs of EEG features, the combination of β band signals (EEG5,6) yields the best performance, outperforming θ band and alpha band signals. Our analysis reveals a strong association between β, γ rhythms and AKE from Microblogs. Previous research has linked low-frequency β rhythms to cognitive activities like attention control, working memory, cognitive control and decision-making, which explains their significant improvement in AKE tasks. High-frequency β rhythms are primarily associated with fine motor movements. Subconscious gaze-following with a pen or a mouse during reading may leads to increased high-frequency β activity. In contrast to the early cognitive processing influenced by β band EEG signal, γ band mainly play a role in the execution and response stages of cognitive activities. As this experiment was conducted during unconscious natural reading, advanced cognitive processing activities were not triggered, resulting in relatively lower levels of γ rhythm.

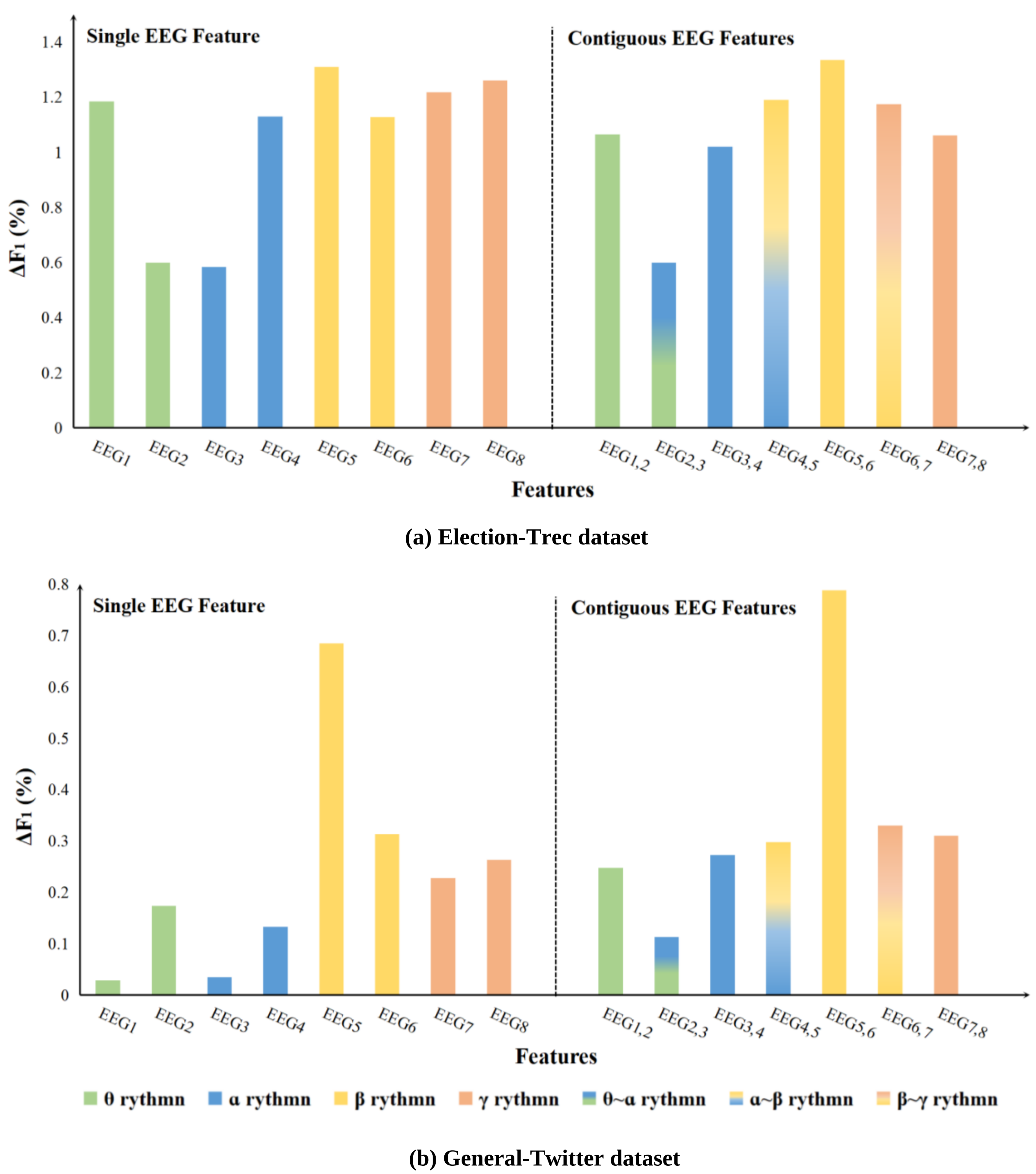

**Fig. 4. $\Delta F_1$ of EEG signals in different rhythms from two datasets**

**Note:** $\Delta F_1$ denotes the cumulative average of the differences between the $F_1$ scores of various models under each feature combination obtained during each training round and the baseline $F_1$ score.

We also conducted AKE from Microblogs tests using eye-tracking signals, and the results demonstrated that the Mean Fixation Duration (MFD) and Total Fixation Duration (TFD) were the two eye-tracking features that exhibited the most prominent performance among the 17 features.

### 4.4. Joint AKE Performance Test of the best-performing EEG signals and Eye-Tracking signals

In addition to combining individual cognitive features, we also analyzed the impact of cross-combining EEG signals and eye-tracking signals on AKE from Microblogs (e.g., Table 6). In this test, we performed cross-combination experiments using the top-performing MFD and TFD features from eye-tracking signals, along with the best-performing β-frequency band features from EEG signals. The control group includes individual cognitive signals, while the combinations of two cognitive signals belong to the experimental group.

| **Feature Combination** | **Election-Trec** | | | | **General-Twitter** | | | |
|---|---|---|---|---|---|---|---|---|
| | Baseline* | ATT-BiLSTM | BiLSTM +CRF | SATT-BiLSTM+CRF | Baseline* | ATT-BiLSTM | BiLSTM +CRF | SATT-BiLSTM +CRF |
| - | 61.67 | 62.13 | 64.64 | 65.38 | 70.67 | 78.64 | 81.63 | 81.74 |
| EEG5 | 62.53 | 62.51 | 65.71 | **67.83** | 74.84 | 78.95 | **81.89** | 82.41 |
| EEG6 | 62.15 | 62.23 | 65.44 | 67.66 | 74.53 | 78.91 | 81.72 | 82.36 |
| ET5(MFD) | 61.91 | 62.05 | 64.7 | 66.34 | 74.55 | 78.94 | 81.3 | 82.24 |
| ET6(TFD) | 62.01 | 62.14 | 64.94 | 66.5 | 73.02 | 78.65 | 81.66 | 82.35 |
| EEG5,6 | **62.57** | **62.59** | 65.62 | 67.77 | 74.71 | **79.27** | 81.83 | **82.49** |
| ET5,6 | 62.05 | 62.12 | 65.59 | 66.75 | 74.34 | 78.83 | 81.13 | 82.39 |
| ET5+EEG5 | 62 | 62.28 | 65.17 | 66.72 | 74.88 | 78.72 | 81.22 | 82.24 |
| ET5+EEG6 | 62.06 | 62.15 | 65.61 | 66.27 | **75.2** | 79.07 | 81.21 | 82.26 |
| ET6+EEG5 | 62.19 | 62.12 | **65.95** | 66.52 | 74.3 | 78.90 | 81.21 | 82.34 |
| ET6+EEG6 | 62.11 | 62.23 | 65.32 | 66.77 | 74.70 | 78.76 | 81.07 | 82.06 |
| ET5+EEG5,6 | 62.07 | 62.17 | 65.48 | 66.56 | 75.1 | 78.62 | 81.28 | 82.04 |
| ET6+EEG5,6 | 62.27 | 62.17 | 65.21 | 66.47 | 74.92 | 78.95 | 81.08 | 82 |
| ET5,6+EEG5 | 62.19 | 62.26 | 65.44 | 66.37 | 74.96 | 78.7 | 81.23 | 82.06 |
| ET5,6+EEG6 | 62.3 | 62.33 | 65.43 | 66.52 | 74.99 | 78.81 | 81.12 | 81.93 |
| ET5,6+EEG5,6 | 62.37 | 62.41 | 65.64 | 66.35 | 74.92 | 78.84 | 81.14 | 82.07 |

**Table 6. $F_1$ Value of AKE model augmented by EEG signals and Eye-Tracking signals**

The $\Delta F_1$ of each combinations of EEG signals and eye-tracking signals are plotted in Fig. 5. We compared the average improvement effects of each feature combination on both datasets.

The combination of β band EEG signals (EEG5,6) demonstrated the best performance on both datasets. On the Election-Trec dataset, the low-frequency β (EEG5) and the combination of Mean Fixation Duration and β band EEG signals (ET5+EEG5,6) performed the next best. On the General-Twitter dataset, the combination of the mean fixation duration and low-frequency beta EEG signals (ET5+EEG5,6) performed the next best, followed by the individual combinations of low-frequency β (EEG5) and high-frequency β (EEG6) EEG signals in descending order of improvement effect.

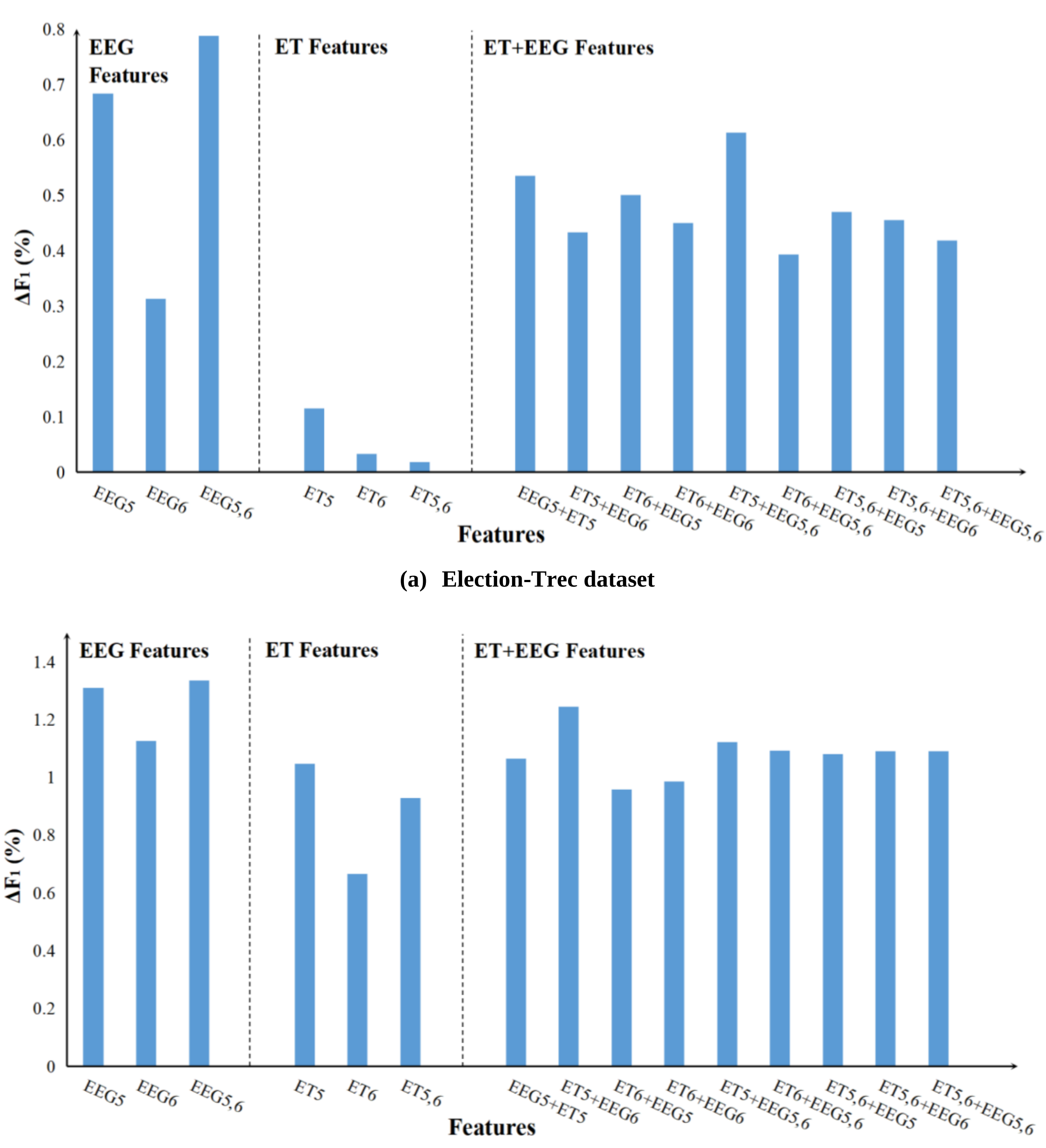

**(a) Election-Trec dataset**

**(b) General-Twitter dataset**

**Fig. 5. $\Delta F_1$ of EEG and eye-tracking signals from two datasets**

The combination of eye-tracking features and EEG features resulted in a modest improvement in the model, but not a significant one. This may be influenced by various factors, including physiological mechanisms, model selection, and OOV words.

First, physiologically, the differential effects of EEG signals and eye-tracking signals on the AKE task may arise from the distinct physiological structures and functions of the eyes and the brain. EEG signals reflect rapid changes in the brain's neuronal electrical activity, making them more suitable for studying sequential text processing features like word boundary recognition, part-of-speech tagging and syntax analysis. Conversely, eye-tracking signals primarily reflect visual processing during reading and fixation, making them more suitable for studying visual text processing.

Second, model selection also plays a significant role in comparing different combinations of cognitive signals generated during human reading. EEG signals, capturing sequential features and contextual information, are better suited for sequence-based methods like RNN, LSTM and Transformer. Eye-tracking signals, reflecting visual attention and reading strategies, are more compatible with attention-based methods. Therefore, the choice of model should consider data volume, feature quality and complexity to avoid overfitting and model instability.

Third, a high proportion of OOV words can impact the experimental results. The significant missing rate of cognitive features likely hinders a complete differentiation of the distinct effects of EEG signals and eye-tracking signals on AKE from Microblogs.

### 4.5. Improved AKE Model based on PLMs

In the following, we summarized the performance of the improved models on the General-Twitter dataset and the Election-Trec dataset (e.g., Table 7). The results highlight the T5-Large model's superiority, achieving the highest performance across all tests. Additionally, the T5-Large model exhibits the most significant enhancement effect of EEG signals on AKE from Microblogs.

Based on the model performances, comparing SATT-BiLSTM+CRF with four pre-trained models, we observed that the T5-Large model achieved the most significant improvement on AKE from Microblogs, followed by the T5-Base model. The BERT model and SATT-BiLSTM+CRF+Glove model showed lower $F_1$ score improvements for AKE, ranging from 3% to 4% compared to the T5 model. The dataset's performance indicates a more pronounced improvement with pre-trained models on small-sample datasets. The $F_1$ score improvement on the Election-Trec dataset is over two times higher than that of the larger General-Twitter dataset. However, the BERT model's performance on the General-Twitter dataset is unsatisfactory, showing lower improvement compared to the SATT-BiLSTM+CRF model.

| Model | Feature Combination | Election-Trec | | | General-Twitter | | |
|---|---|---|---|---|---|---|---|
| | | P | R | $F_1$ | P | R | $F_1$ |
| SATT-BiLSTM +CRF | - | 81.84 | 66.83 | 65.38 | 86.18 | 80.85 | 81.74 |
| | All_EEG | 72.6 | 65.87 | 67.82 | 84.72 | 81.61 | 82.46 |
| | All_ET | 72.71 | 65.77 | 67.4 | 84.53 | 81.61 | 82.22 |
| | All_ET& All_EEG | 72.03 | 65.57 | 67.66 | 84.6 | 81.73 | 82.26 |
| SATT-BiLSTM +CRF+Glove | - | 72.16 | **67.41** | 69.7 | 88.21 | 78.77 | 82.7 |
| | All_EEG | 75.29 | 67.63 | 71.26 | 85.51 | 83.03 | 83.89 |
| | All_ET | 76.78 | 64.99 | 70.4 | 85.42 | 83.1 | 83.66 |
| | All_ET& All_EEG | 76.7 | 65.45 | 70.63 | 85.23 | 82.24 | 83.71 |
| BERT | - | 72.11 | **71.47** | 71.79 | 80.97 | 81.9 | 81.43 |
| | All_EEG | 74.62 | 71.17 | 72.85 | 80.84 | 82.61 | 81.71 |
| | All_ET | 71.48 | 70.58 | 71.03 | 81.77 | 81.36 | 81.56 |
| | All_ET& All_EEG | 72.71 | 70.96 | 71.82 | 81.41 | 82.11 | 81.76 |
| T5-Base | - | 85.52 | 66.34 | 74.72 | 90.14 | 79.05 | 84.23 |
| | All_EEG | 85.21 | 69.58 | 76.61 | 91.1 | 78.5 | 84.33 |
| | All_ET | 84.29 | 68.32 | 75.47 | 92.68 | 78.16 | 84.8 |
| | All_ET& All_EEG | 86.68 | 66.25 | 75.1 | 91.8 | 78.3 | 84.51 |
| T5-Large | - | 86.68 | 66.25 | 75.1 | 91.49 | 78.43 | 84.46 |
| | All_EEG | 85.95 | 68.92 | 76.5 | 92.89 | 79.55 | **85.7** |
| | All_ET | 84.23 | 67.07 | 74.68 | 91.33 | 78.99 | 84.72 |
| | All_ET& All_EEG | **87.04** | 69.02 | **76.99** | **93.13** | 79.37 | **85.7** |

**Table 7. $F_1$ value of Improved AKE based on pre-trained models**

Fig. 6 compares the performance of three improvement strategies on two datasets. EEG signals have a more significant impact on AKE from Microblogs than eye-tracking signals. The combination of “ET+EEG” performs the best under the T5-Large model, achieving the highest improvement among all cognitive feature combinations. In summary, the T5-Large model excels in learning representations of cognitive features and semantic understanding of Microblog text, demonstrating the best performance. The BERT and T5-Base models show relatively better performance on small-scale datasets, but their improvement effects for different feature combinations on the General-Twitter dataset are relatively weak and similar, making further fine-grained feature analysis challenging. This phenomenon is even more pronounced on the larger General-Twitter dataset, possibly due to the model's learned textual features overshadowing the influence of cognitive features on the task's performance.

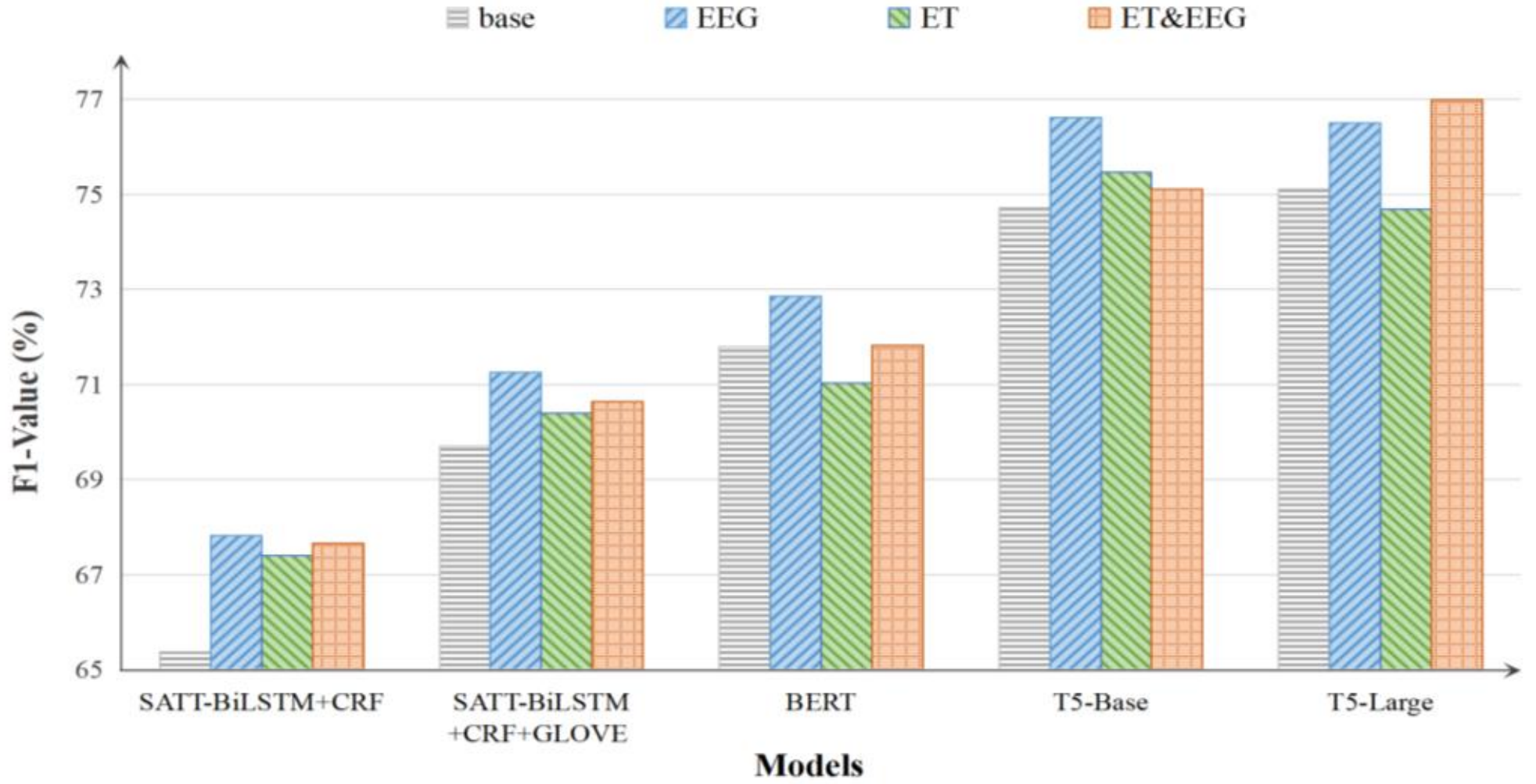

**(a) Election-Trec Dataset**

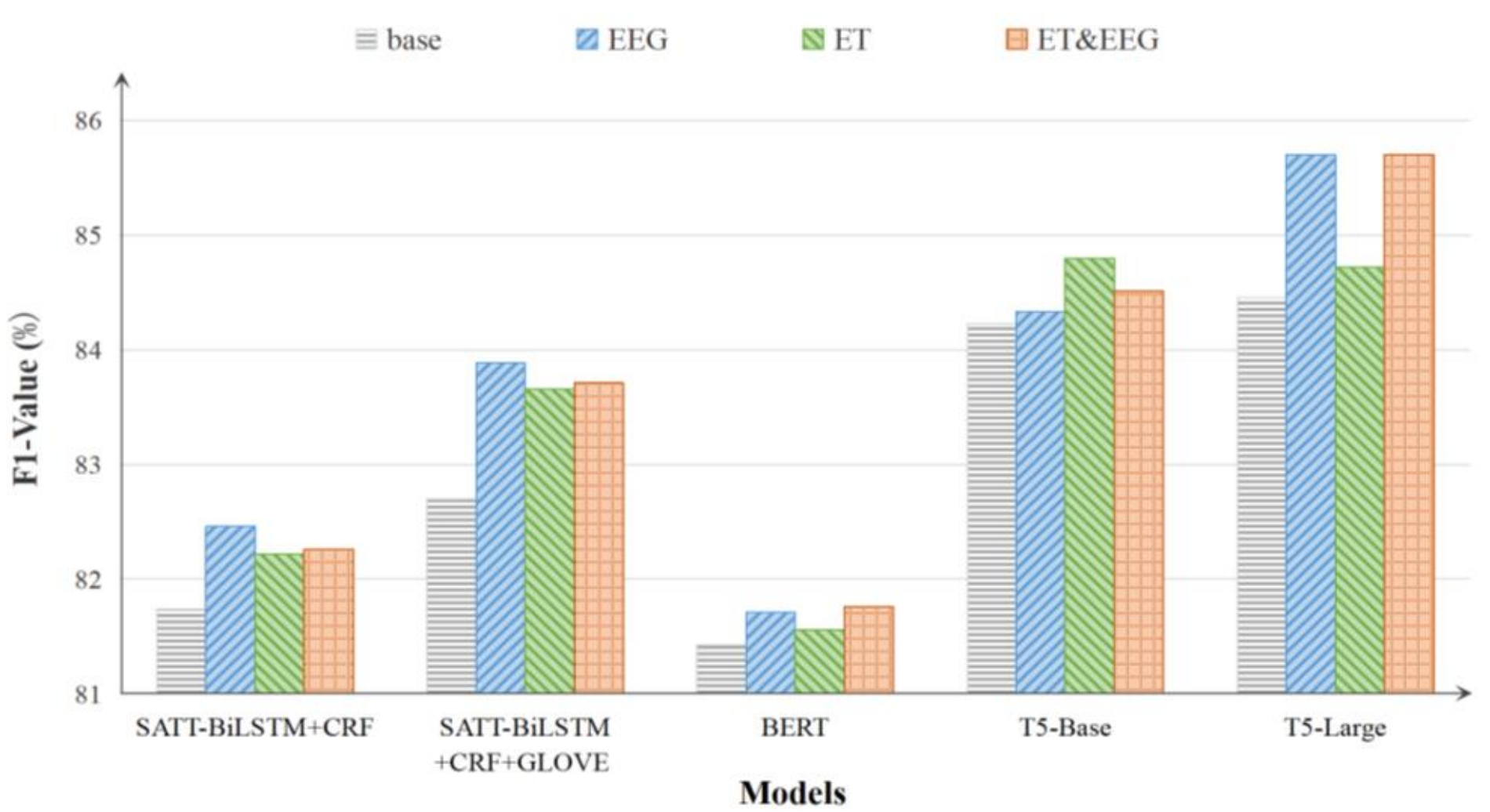

**(b) General-Twitter dataset**

**Fig. 6. $F_1$ values of improvement models form two datasets**

In our experiment, sentence sequences were first input into the model, and then combining their output predictions with cognitive features. However, the pre-trained model's excellent semantic comprehension of the input utterances diminishes the significance of cognitive signals as external features. The T5-Large model stands as an exception, its exceptional language modeling and generalization capabilities allow it to express more language features compared to the T5-Base model, making it exceptionally well-suited for new domains or tasks.

### 4.5. Case Study

| Features | Election-Trec | General-Twitter |
|---|---|---|
| - | **Ele1:** I know a ***basket of deplorables*** that won't shed a tear when you die 71% of 250 physicians polled say Hillary ...URL<br>**Ele2:** The gender wage gap is even wider for women of color.It's time to ensure equal pay. ***Black womens equal pay*** URL<br>**Ele3:** make no mistake, the world is watching our ***2016 election*** and it's an iq test we might not pass.<br>**Ele4:** ***Brexit*** may mean Britons have to have visas - not a chance even the hardest ***brexit*** will have uk url eu visa-free travel.<br>**Ele5:** trump said at the second ***debate*** that he disagrees with his own running mate on the issue of Syria URL | **Gen1:** have an ***awesome*** week~mike"<br>**Gen2:** Los y las fans de ***norman tuiteamo norman reedus en comic con chile*** NAME NAME NAME<br>**Gen3:** Mtnan jme suis lever jdanse le galsen tu peu pas teste no muzik ***team221*** samadineubi thiakhagoune mbolokhe mbolokhe<br>**Gen4:** Obvio que ganara la fernanda gallardo es la major de ***calle7*** aguante<br>**Gen5:** More men died in last year through suicide than in all wars since 1945 we need tomorrow's ***world suicide prevention day*** more than ever |
| ALL_EEG | **Ele1:** I know a ***basket of deplorables*** that won't shed a tear when you die 71% of 250 physicians polled say Hillary ...URL<br>**Ele2:** The gender wage gap is even wider for women of color.It's time to ensure equal pay. ***Black womens equal pay*** URL<br>**Ele3:** make no mistake, the world is watching our ***2016 election*** and it's an iq test we might not pass.<br>**Ele4:** ***Brexit*** may mean Britons have to have visas - not a chance even the hardest ***brexit*** will have uk url eu visa-free travel.<br>**Ele5:** trump said at the second ***debate*** that he disagrees with his own running mate on the issue of Syria URL | **Gen1:** have an ***awesome*** week~mike"<br>**Gen2:** Los y las fans de ***norman tuiteamo norman reedus en comic con chile*** NAME NAME NAME<br>**Gen3:** Mtnan jme suis lever jdanse le galsen tu peu pas teste no muzik ***team221*** samadineubi thiakhagoune mbolokhe mbolokhe<br>**Gen4:** Obvio que ganara la fernanda Gallardo es la mejor de ***calle7*** aguante<br>**Gen5:** More men died in last year through suicide than in all wars since 1945 we need tomorrow's ***world suicide prevention day*** more than ever |
| ALL_ET | **Ele1:** I know a ***basket of deplorables*** that won't shed a tear when you die 71% of 250 physicians polled say Hillary ...URL<br>**Ele2:** The gender wage gap is even wider for women of color.It's time to ensure equal pay. ***Black womens equal pay*** URL<br>**Ele3:** make no mistake, the world is watching our ***2016 election*** and it's an iq test we might not pass.<br>**Ele4:** ***Brexit*** may mean Britons have to have visas - not a chance even the hardest ***brexit*** will have uk url eu visa-free travel.<br>**Ele5:** trump said at the second ***debate*** that he disagrees with his own running mate on the issue of Syria URL | **Gen1:** have an ***awesome*** week~mike"<br>**Gen2:** Los y las fans de ***norman tuiteamo norman reedus en comic con chile*** NAME NAME NAME<br>**Gen3:** Mtnan jme suis lever jdanse le galsen tu peu pas teste no muzik ***team221*** samadineubi thiakhagoune mbolokhe mbolokhe<br>**Gen4:** Obvio que ganara la fernanda gallardo es la major de ***calle7*** aguante<br>**Gen5:** More men died in last year through suicide than in all wars since 1945 we need tomorrow's ***world suicide prevention day*** more than ever |
| ALL_EEG & ALL_ET | **Ele1:** I know a ***basket of deplorables*** that won't shed a tear when you die 71% of 250 physicians polled say Hillary ...URL<br>**Ele2:** The gender wage gap is even wider for women of color.It's time to ensure equal pay. ***Black womens equal pay*** URL<br>**Ele3:** make no mistake, the world is watching our ***2016 election*** and it's an iq test we might not pass.<br>**Ele4:** ***Brexit*** may mean Britons have to have visas - not a chance even the hardest ***brexit*** will have uk url eu visa-free travel.<br>**Ele5:** trump said at the second ***debate*** that he disagrees with his own running mate on the issue of Syria URL | **Gen1:** have an ***awesome*** week~mike"<br>**Gen2:** Los y las fans de ***norman tuiteamo norman reedus en comic con chile*** NAME NAME NAME<br>**Gen3:** Mtnan jme suis lever jdanse le galsen tu peu pas teste no muzik ***team221*** samadineubi thiakhagoune mbolokhe mbolokhe<br>**Gen4:** Obvio que ganara la fernanda gallardo es la major de ***calle7*** aguante<br>**Gen5:** More men died in last year through suicide than in all wars since 1945 we need tomorrow's ***world suicide prevention day*** more than ever |

**Note**: *Bold italicized mark* indicate manually annotated correct Hashtags, *blue mark* represent correctly predicted keyphrases, *green mark* indicate predicted incorrect results, *yellow mark* represent partially predicted words for the target answers.

**Table 8. Example of AKE incorporating Cognitive Signals Generated during Human Reading**

In this section, we randomly selected five instances from the Election-Trec dataset and the General-Twitter dataset to visually illustrate the impact of cognitive signals generated during human reading on AKE from Microblogs.

As presented in Table 8, we use different colors and fonts to distinguish the final predictions. The results showed that when EEG signals were combined, AKE from Microblogs obtained Hashtags that were most similar to the correct answers, with only 2 instances not being entirely predicted correctly. The predictions based on eye-tracking signals and EEG signals performed better on the Election-Trec dataset, with only Ele5 not being predicted correctly. Comparing the two cognitive signals, the enhancement effect of eye-tracking signals on AKE was significantly less pronounced than that of EEG signals, with their prediction results being similar to the baseline. This once again confirms the conclusion obtained from the above experiments: EEG signals have a remarkable enhancing effect on AKE from Microblogs.

In order to compare the evaluation results more intuitively, we used the following scoring criteria: 10 points for correct predictions, 3 points for partially correct predictions, and 0 points for incorrect predictions. The scores for each feature combination are as follow: "-: 12 points, ALL_EEG: 86 points, ALL_ET: 29 points, and ALL_ET& ALL_EEG: 53 points". These results clearly indicate that cognitive signals generated during human reading have a positive impact on the AKE from Microblogs. Among them, EEG signals show a stronger enhancement on AKE performance, while eye-tracking signals exhibit a relatively weaker enhancing capability.

This case study demonstrates that incorporating EEG signals has a remarkable enhancing effect on AKE from Microblogs, resulting in predictions that are most similar to correct answers. While eye-tracking signals also contribute to AKE performance, their enhancement effect is less pronounced compared to EEG signals.

## 5. Discussion

In this section, we discuss the significance and limitations of our research. In Section 5.1, we analyze our study's value from theoretical and practical perspectives. In Section 5.2, we address existing limitations and propose corresponding improvement suggestions.

### 5.1. Implication

In the preceding section, we confirmed the enhancing effect of cognitive signals, particularly EEG signals, on the AKE from Microblogs task through various tests. Furthermore, in the improved model, cognitive signals demonstrated exceptional performance on the T5-Large model, surpassing previous results. In this subsection, we will present a detailed analysis of the theoretical and practical significance of the research findings in this paper.

**(1) Theoretical Implications**

Our research enriches the theoretical foundation of AKE by revealing the cognitive psychology's role as a theoretical support for this task. AKE, being an NLP task involving semantic understanding and crucial information identification, requires a profound understanding of human cognition and linguistic information processing. Therefore, our study incorporates relevant theories and experimental research from cognitive psychology to explore the processing rules and mechanisms of human text comprehension, leading to advancements in theoretical models of language processing. Additionally, the research outcomes from AKE can provide experimental design and data analysis support for NLP tasks with biological signals, promoting the interdisciplinary integration of both fields.

Moreover, our research provides theoretical support for the development of novel AKE methods based on biological signals, leading to more efficient and accurate NLP techniques. Through the analysis of cognitive signals during human readings, we can gain profound insights into how humans comprehend language, thereby advancing research in NLP. The processing and integration of cognitive signals generated during human reading for AKE tasks rely on support from diverse interdisciplinary fields, including computer science, psychology, and neuroscience. As a result, this interdisciplinary research fosters communication and collaboration among different fields, facilitating their integration and development.

Additionally, the integration of cognitive signals in AKE tasks contributes to the exploration of multimodal NLP approaches. Multimodal NLP involves the fusion and analysis of different modalities such as text, images, speech, and gestures to improve natural language understanding and generation. The inclusion of cognitive signals adds an additional modality to this research domain, enabling the investigation of how cognitive processes interact with other modalities during language comprehension. This expands the scope of multimodal NLP research and facilitates the development of more comprehensive and robust models.

**(2) Practical Implications**

Our research offers practical implications for optimizing the performance of AKE and other NLP tasks, providing valuable insights into the development of related research methods and experimental data. By incorporating cognitive signals generated during human reading, the model gains access to richer external information, allowing it to simulate human readings mechanisms for better text analysis. Additionally, using an unstructured Microblog short text corpus for AKE enhances the robustness and adaptability of experiments to various reading scenarios and readers. This establishes a reliable and broadly applicable practical foundation for future advancements in AKE research.

In some sense, our research expands the application scenarios of cognitive signals in NLP. Due to current technological and resource limitations, general corpora with cognitive signals are scarce, mainly focused on areas like sports impact, sentiment analysis, and rehabilitation therapy. This limited availability of corpora creates a disciplinary bias in the vocabulary distribution related to cognitive signals, significantly affecting the performance of AKE from Microblogs. To address this, our research assigns average corpus features to OOV words related to cognitive signals and improves the combination position and model selection of cognitive signals to better integrate them into the model. Only when the model fully learns the features provided by cognitive signals can we objectively assess their impact on AKE from Microblogs and derive meaningful reference values from the experimental results.

AKE service with high accuracy and efficiency can aid researchers in conducting research quickly, especially in non-specialized fields. Scholars often use keywords from academic papers to characterize specific research topics and study their evolution using bibliometric methods like word frequency and co-word analysis. However, incomplete keyword information can hinder non-specialized readers' understanding of semantic relationships among these keywords. By incorporating cognitive signals into AKE, we can provide more comprehensive semantic information in keywords, enabling readers to gain a comprehensive and accurate understanding of the subject in the field. Furthermore, this approach can showcase the evolution of research topics in that discipline through keywords.

### 5.2. Limitation

Our study also has some limitations. The presence of a high proportion of OOV words in the AKE dataset poses a significant challenge to the integration of cognitive features. The ZUCO dataset contains less than five thousand vocabulary words, while the smaller AKE dataset has a vocabulary size of 37,347, indicating a high proportion of OOV words. We have considered several approaches to address this problem: 1) Calculate semantic similarity between OOV words and existing words and use the similarity as a weight to compute the feature values of each OOV word. 2) Utilize external cognitive signal corpora to map features to the word level and align them with OOV words. 3) Reconstruct the corpus based on character-level features, using character n-grams (e.g., character n-gram frequency) or character embeddings as feature values. Character-level features of OOV words can be learned from training datasets or other external resources.

Additionally, in the test of Section 4.4, we only selected the top two performing features from each modality for the combination experiment, thus not covering all potential combinations. Due to equipment and time constraints, we couldn't test all combinations in this study. Future research could explore other combinations of eye-tracking signals and EEG signals, potentially leading to further discoveries.

From the perspective of model selection, we only adopts a relatively simple approach, incorporating attention mechanisms and CRF on top of BiLSTM. While improvement experiments using pre-trained models were proposed in Section 4.5, there was no detailed investigation into the influence of each combination of eye-tracking signals and EEG signals on AKE from Microblogs performance. Subsequent research could investigate the performance of cognitive signals in AKE using Transformer, CNN, and other pre-trained models.

From the perspective of feature selection, our study solely used eye-tracking signals and EEG signals as cognitive signals, both derived from the same dataset. Ideally, incorporating diverse cognitive signals from various open-source datasets, processing and unifying them into a single dimension, followed by AKE from Microblogs test, would yield more convincing experimental results and potentially further enhance AKE from Microblogs performance.

## 6. Conclusions and Future Work

In this study, we integrated EEG signals and eye-tracking signals, generated during human reading, as external features into AKE from Microblogs. Improvements were made in both feature combination and model structure to optimize the performance of models. In feature combinations, we extensively explored cross-combinations of eye-tracking signals and EEG signals, conducting comprehensive tests their performance on AKE from Microblogs. As for models, we not only added CRF and two types of attention mechanisms to Bi-LSTM, but also further improved the model based on LLMs. The experimental results show that EEG signals significantly enhance AKE from Microblogs, especially the low-frequency β rhythm, which has the most notable impact on the model. Combining the best-performing eye-tracking signals with EEG signals resulted in performance between the two separate signal experiments. This variation could be due to differences in signal-to-noise ratios between eye-tracking signals and EEG signals, reflecting distinct capabilities in capturing cognitive activities during human reading.

This study's significance extends beyond AKE from Microblogs. Integrating cognitive signals like EEG and eye-tracking helps us understand the complexity of human reading behavior comprehensively. EEG signals reveal fundamental neural processes in reading comprehension, while eye-tracking signals depict individual cognitive activities during reading. By exploring their associations, we uncover how attention, comprehension, and memory synergistically contribute to reading. This deepens our understanding of the neural mechanisms in reading comprehension, information processing, and learning, providing valuable research foundations in education and cognitive science.

In the future, this research can expand both horizontally and vertically. Horizontally, cognitive signals generated during human reading like can be applied to other NLP tasks such as text generation, speech recognition, and machine translation. Vertically, AKE model's performance can be improved through corpus replacement, model optimization and enhancing cognitive signals. In order to explore human reading behavior more deeply and leverage user cognitive signals more effectively in NLP tasks, we should focus on the prospects below. Firstly, through exploring multi-sources cognitive signals (EEG, eye-tracking, MRI and others) generated during human reading, we can gain a comprehensive understanding of cognitive information. Secondly, cross-modal learning, combining various cognitive signals with textual information, can capture correlations between modalities and enhance NLP performance. Building datasets that record cognitive signals of human reading behavior is crucial. Larger, diverse datasets covering different texts, so that researchers are able to further understand the nature of human reading behavior and its connection with NLP tasks. Besides, translating theoretical research into practical applications is also vital for the study's prospects. Personalized NLP applications based on user-specific cognitive signals can enhance user experience and effectiveness. In summary, as we delve deeper into the study of human reading behavior and the integration of cognitive signals, NLP technology is poised to advance towards more intelligent and user-centric language processing applications.

## ACKNOWLEDGMENTS

This work is supported by National Natural Science Foundation of China (Grant No.72074113).

## ORCID

Chengzhi Zhang： https://orcid.org/0000-0001-9522-2914

## ENDNOTES

1. https://osf.io/2urht/#!
2. https://www.eyetracking-eeg.org/
3. http://qizhang.info/paper/
4. https://trec.nist.gov/data/tweets/

## REFERENCES

Bahdanau, D., Cho, K., & Bengio, Y. (2016). *Neural Machine Translation by Jointly Learning to Align and Translate* (arXiv:1409.0473). arXiv. https://doi.org/10.48550/arXiv.1409.0473

Bao, T., & Zhang, C. (2023). Extracting Chinese Information with ChatGPT：An Empirical Study by Three Typical Tasks. *Data Analysis and Knowledge Discovery*, *7*(9), 1–11. https://doi.org/10.11925/infotech.2096-3467.2023.0473

Barrett, M., Bingel, J., Keller, F., & Søgaard, A. (2016). Weakly Supervised Part-of-speech Tagging Using Eye-tracking Data. *Proceedings of the 54th Annual Meeting of the Association for Computational Linguistics (Volume 2: Short Papers)*, 579–584. https://doi.org/10.18653/v1/P16-2094

Bellaachia, A., & Al-Dhelaan, M. (2012). NE-Rank: A Novel Graph-Based Keyphrase Extraction in Twitter. *2012 IEEE/WIC/ACM International Conferences on Web Intelligence and Intelligent Agent Technology*, *1*, 372–379. https://doi.org/10.1109/WI-IAT.2012.82

Bennani-Smires, K., Musat, C., Hossmann, A., Baeriswyl, M., & Jaggi, M. (2018). Simple Unsupervised Keyphrase Extraction using Sentence Embeddings. *Proceedings of the 22nd Conference on Computational Natural Language Learning*, 221–229. https://doi.org/10.18653/v1/K18-1022

Boudin, F. (2018). Unsupervised Keyphrase Extraction with Multipartite Graphs. *Proceedings of the 2018 Conference of the North American Chapter of the Association for Computational Linguistics: Human Language Technologies, Volume 2 (Short Papers)*, 667–672. https://doi.org/10.18653/v1/N18-2105

Bougouin, A., Boudin, F., & Daille, B. (2013). TopicRank: Graph-Based Topic Ranking for Keyphrase Extraction. *Proceedings of the Sixth International Joint Conference on Natural Language Processing*, 543–551. https://aclanthology.org/I13-1062

Bueno, A. P. A., Sato, J. R., & Hornberger, M. (2019). Eye tracking—The overlooked method to measure cognition in neurodegeneration? *Neuropsychologia*, *133*, 107191. https://doi.org/10.1016/j.neuropsychologia.2019.107191

Campos, R., Mangaravite, V., Pasquali, A., Jorge, A. M., Nunes, C., & Jatowt, A. (2018). A Text Feature Based Automatic Keyword Extraction Method for Single Documents. In G. Pasi, B. Piwowarski, L. Azzopardi, & A. Hanbury (Eds.), *Advances in Information Retrieval* (pp. 684–691). Springer International Publishing.

Chin, J. Y., Bhowmick, S. S., & Jatowt, A. (2019). On-demand recent personal tweets summarization on mobile devices. *Journal of the Association for Information Science and Technology*, *70*(6), 547–562. https://doi.org/10.1002/asi.24137

Chung, H. W., Hou, L., Longpre, S., Zoph, B., Tay, Y., Fedus, W., Li, Y., Wang, X., Dehghani, M., Brahma, S., Webson, A., Gu, S. S., Dai, Z., Suzgun, M., Chen, X., Chowdhery, A., Castro-Ros, A., Pellat, M., Robinson, K., … Wei, J. (2022). *Scaling Instruction-Finetuned Language Models* (arXiv:2210.11416; Version 5). arXiv. https://doi.org/10.48550/arXiv.2210.11416

Daza, R., Morales, A., Fierrez, J., & Tolosana, R. (2021). mEBAL: A Multimodal Database for Eye Blink Detection and Attention Level Estimation. *Companion Publication of the 2020 International Conference on Multimodal Interaction*, 32–36. https://doi.org/10.1145/3395035.3425257

de León Rodríguez, D., Mouthon, M., Annoni, J.-M., & Khateb, A. (2022). Current exposure to a second language modulates bilingual visual word recognition: An EEG study. *Neuropsychologia*, *164*, 108109. https://doi.org/10.1016/j.neuropsychologia.2021.108109

De Vito, D., Ferrey, A. E., Fenske, M. J., & Al-Aidroos, N. (2018). Cognitive-behavioral and electrophysiological evidence of the affective consequences of ignoring stimulus representations in working memory. *Cognitive, Affective, & Behavioral Neuroscience*, *18*(3), 460–475. https://doi.org/10.3758/s13415-018-0580-x

Demirel, S., De Moraes, C. G. V., Gardiner, S. K., Liebmann, J. M., Cioffi, G. A., Ritch, R., Gordon, M. O., & Kass, M. A. (2012). The Rate of Visual Field Change in the Ocular Hypertension Treatment Study. *Investigative Ophthalmology & Visual Science*, *53*(1), 224–227. https://doi.org/10.1167/iovs.10-7117

Dudschig, C. (2022). Language and non-linguistic cognition: Shared mechanisms and principles reflected in the N400. *Biological Psychology*, *169*, 108282. https://doi.org/10.1016/j.biopsycho.2022.108282

Dutt, R., Hiware, K., Ghosh, A., & Bhaskaran, R. (2018). SAVITR: A System for Real-time Location Extraction from Microblogs during Emergencies. *Companion of the The Web Conference 2018 on The Web Conference 2018 - WWW '18*, 1643–1649. https://doi.org/10.1145/3184558.3191623

El-Beltagy, S. R., & Rafea, A. (2009). KP-Miner: A keyphrase extraction system for English and Arabic documents. *Information Systems*, *34*(1), 132–144. https://doi.org/10.1016/j.is.2008.05.002

Funke, G., Greenlee, E., Carter, M., Dukes, A., Brown, R., & Menke, L. (2016). Which Eye Tracker Is Right for Your Research? Performance Evaluation of Several Cost Variant Eye Trackers. *Proceedings of the Human Factors and Ergonomics Society Annual Meeting*, *60*(1), 1240–1244. https://doi.org/10.1177/1541931213601289

Fushimi, T., & Kanno, K. (2020). Extraction of user demands based on similar tweets graph. *Proceedings of the 2019 IEEE/ACM International Conference on Advances in Social Networks Analysis and Mining*, 1005–1012. https://doi.org/10.1145/3341161.3344824

Ganushchak, L., Christoffels, I., & Schiller, N. (2011). The Use of Electroencephalography in Language Production Research: A Review. *Frontiers in Psychology*, *2*. https://www.frontiersin.org/articles/10.3389/fpsyg.2011.00208

Gaonkar, S., Li, J., Choudhury, R. R., Cox, L., & Schmidt, A. (2008). Micro-Blog: Sharing and querying content through mobile phones and social participation. *Proceedings of the 6th International Conference on Mobile Systems, Applications, and Services*, 174–186. https://doi.org/10.1145/1378600.1378620

Gavaret, M., Iftimovici, A., & Pruvost-Robieux, E. (2023). EEG: Current relevance and promising quantitative analyses. *Revue Neurologique*, *179*(4), 352–360. https://doi.org/10.1016/j.neurol.2022.12.008

Geraets, C. N. W., Klein Tuente, S., Lestestuiver, B. P., van Beilen, M., Nijman, S. A., Marsman, J. B. C., & Veling, W. (2021). Virtual reality facial emotion recognition in social environments: An eye-tracking study. *Internet Interventions*, *25*, 100432. https://doi.org/10.1016/j.invent.2021.100432

Gkikas, D. C., Tzafilkou, K., Theodoridis, P. K., Garmpis, A., & Gkikas, M. C. (2022). How do text characteristics impact user engagement in social media posts: Modeling content readability, length, and hashtags number in Facebook. *International Journal of Information Management Data Insights*, *2*(1), 100067. https://doi.org/10.1016/j.jjimei.2022.100067

Gong, Y., Zhang, Q., & Huang, X. (2018). Hashtag recommendation for multimodal microblog posts. *Neurocomputing*, *272*, 170–177. https://doi.org/10.1016/j.neucom.2017.06.056

Goz, F., & Mutlu, A. (2023). SkyWords: An automatic keyword extraction system based on the skyline operator and semantic similarity. *Engineering Applications of Artificial Intelligence*, *123*, 106338. https://doi.org/10.1016/j.engappai.2023.106338

Gupta, P., Chugh, K., Dhall, A., & Subramanian, R. (2020). The eyes know it: FakeET- An Eye-tracking Database to Understand Deepfake Perception. *Proceedings of the 2020 International Conference on Multimodal Interaction*, 519–527. https://doi.org/10.1145/3382507.3418857

Hassani, H., Ershadi, M. J., & Mohebi, A. (2022). LVTIA: A new method for keyphrase extraction from scientific video lectures. *Information Processing & Management*, *59*(2), 102802. https://doi.org/10.1016/j.ipm.2021.102802

Hollenstein, N., Barrett, M., & Beinborn, L. (2020). Towards Best Practices for Leveraging Human Language Processing Signals for Natural Language Processing. *Proceedings of the Second Workshop on Linguistic and Neurocognitive Resources*, 15–27. https://aclanthology.org/2020.lincr-1.3

Hollenstein, N., Barrett, M., Troendle, M., Bigiolli, F., Langer, N., & Zhang, C. (2019a). *Advancing NLP with Cognitive Language Processing Signals*. http://doc.paperpass.com/patent/arXiv190402682.html

Hollenstein, N., Barrett, M., Troendle, M., Bigiolli, F., Langer, N., & Zhang, C. (2019b). *Advancing NLP with Cognitive Language Processing Signals* (arXiv:1904.02682; Version 1). arXiv. https://doi.org/10.48550/arXiv.1904.02682

Hollenstein, N., Rotsztejn, J., Troendle, M., Pedroni, A., Zhang, C., & Langer, N. (2018). ZuCo, a simultaneous EEG and eye-tracking resource for natural sentence reading. *Scientific Data*, *5*(1), 180291. https://doi.org/10.1038/sdata.2018.291

Hu, Y., Ameer, I., Zuo, X., Peng, X., Zhou, Y., Li, Z., Li, Y., Li, J., Jiang, X., & Xu, H. (2023). *Zero-shot Clinical Entity Recognition using ChatGPT* (arXiv:2303.16416). arXiv. https://doi.org/10.48550/arXiv.2303.16416

Ikhwantri, F., Putra, J. W. G., Yamada, H., & Tokunaga, T. (2023). Looking deep in the eyes: Investigating interpretation methods for neural models on reading tasks using human eye-movement behaviour. *Information Processing & Management*, *60*(2), 103195. https://doi.org/10.1016/j.ipm.2022.103195

Jeong, D., Oh, S., & Park, E. (2022). DemoHash: Hashtag recommendation based on user demographic information. *Expert Systems with Applications*, *210*, 118375. https://doi.org/10.1016/j.eswa.2022.118375

Jiang, X. (2008). A Keyword Extraction Method Based on Lexical Chains. In L. Kang, Z. Cai, X. Yan, & Y. Liu (Eds.), *Advances in Computation and Intelligence* (pp. 360–367). Springer. https://doi.org/10.1007/978-3-540-92137-0_40

Jin, J., Wang, A., Wang, C., & Ma, Q. (2023). How do consumers perceive and process online overall vs. individual text-based reviews? Behavioral and eye-tracking evidence. *Information & Management*, *60*(5), 103795. https://doi.org/10.1016/j.im.2023.103795

Kang, H., Sun, X., Feng, T., & Lin, S. (2021). Keyword Extraction Based on Semantic Similarity Metric and Multi-Feature Computing. *2021 7th International Conference on Big Data Computing and Communications (BigCom)*, 188–195. https://doi.org/10.1109/BigCom53800.2021.00003

Keles, T., Yildiz, A. M., Barua, P. D., Dogan, S., Baygin, M., Tuncer, T., Demir, C. F., Ciaccio, E. J., & Acharya, U. R. (2023). A new one-dimensional testosterone pattern-based EEG sentence classification method. *Engineering Applications of Artificial Intelligence*, *119*, 105722. https://doi.org/10.1016/j.engappai.2022.105722

Khosla, A., Khandnor, P., & Chand, T. (2020). A comparative analysis of signal processing and classification methods for different applications based on EEG signals. *Biocybernetics and Biomedical Engineering*, *40*(2), 649–690. https://doi.org/10.1016/j.bbe.2020.02.002

Kim, D.-W., & Im, C.-H. (2018). EEG Spectral Analysis. In C.-H. Im (Ed.), *Computational EEG Analysis: Methods and Applications* (pp. 35–53). Springer. https://doi.org/10.1007/978-981-13-0908-3_3

Kok, E. M., & Jarodzka, H. (2017). Before your very eyes: The value and limitations of eye tracking in medical education. *Medical Education*, *51*(1), 114–122. https://doi.org/10.1111/medu.13066

Koloski, B., Pollak, S., Škrlj, B., & Martinc, M. (2022). Out of Thin Air: Is Zero-Shot Cross-Lingual Keyword Detection Better Than Unsupervised? *Proceedings of the Thirteenth Language Resources and Evaluation Conference*, 400–409. https://aclanthology.org/2022.lrec-1.42

Kumar, N., & Srinathan, K. (2008). Automatic keyphrase extraction from scientific documents using N-gram filtration technique. *DocEng '08 Proceeding of the Eighth ACM Symposium on Document Engineering*. https://doi.org/10.1145/1410140.1410180

Lee, W., Chun, M., Jeong, H., & Jung, H. (2023). Toward Keyword Generation through Large Language Models. *Companion Proceedings of the 28th International Conference on Intelligent User Interfaces*, 37–40. https://doi.org/10.1145/3581754.3584126

Li, J., Liao, M., Gao, W., He, Y., & Wong, K.-F. (2016). Topic Extraction from Microblog Posts Using Conversation Structures. *Proceedings of the 54th Annual Meeting of the Association for Computational Linguistics (Volume 1: Long Papers)*, 2114–2123. https://doi.org/10.18653/v1/P16-1199

Li, X., Hu, B., Xu, T., Shen, J., & Ratcliffe, M. (2015). A study on EEG-based brain electrical source of mild depressed subjects. *Computer Methods and Programs in Biomedicine*, *120*(3), 135–141. https://doi.org/10.1016/j.cmpb.2015.04.009

Li, & Zhao, J. (2016). TextRank Algorithm by Exploiting Wikipedia for Short Text Keywords Extraction. *2016 3rd International Conference on Information Science and Control Engineering (ICISCE)*, 683–686. https://doi.org/10.1109/ICISCE.2016.151

Liang, X., Wu, S., Li, M., & Li, Z. (2021). Unsupervised Keyphrase Extraction by Jointly Modeling Local and Global Context. *Proceedings of the 2021 Conference on Empirical Methods in Natural Language Processing*, 155–164. https://doi.org/10.18653/v1/2021.emnlp-main.14

Liu, R., Lin, Z., & Wang, W. (2021). Addressing Extraction and Generation Separately: Keyphrase Prediction With Pre-Trained Language Models. *IEEE/ACM Transactions on Audio, Speech and Language Processing*, *29*, 3180–3191. https://doi.org/10.1109/TASLP.2021.3120587

Lum, J. A. G., Clark, G. M., Bigelow, F. J., & Enticott, P. G. (2022). Resting state electroencephalography (EEG) correlates with children's language skills: Evidence from sentence repetition. *Brain and Language*, *230*, 105137. https://doi.org/10.1016/j.bandl.2022.105137

Lundqvist, M., Brincat, S. L., Rose, J., Warden, M. R., Buschman, T. J., Miller, E. K., & Herman, P. (2023). Working memory control dynamics follow principles of spatial computing. *Nature Communications*, *14*(1), Article 1. https://doi.org/10.1038/s41467-023-36555-4

Messaoudi, C., Guessoum, Z., & Romdhane, L. ben. (2022). A Deep Learning Model for Opinion mining in Twitter Combining Text and Emojis. *Procedia Computer Science*, *207*, 2628–2637. https://doi.org/10.1016/j.procs.2022.09.321

Mihalcea, R., & Tarau, P. (2004). TextRank: Bringing Order into Text. *Proceedings of the 2004 Conference on Empirical Methods in Natural Language Processing*, 404–411. https://aclanthology.org/W04-3252

Mizuka, K., Suzuki, Y., & Nadamoto, A. (2017). Extraction of commentary tweets about news articles. *Proceedings of the 19th International Conference on Information Integration and Web-Based Applications & Services*, 188–192. https://doi.org/10.1145/3151759.3151841

Nasar, Z., Jaffry, S. W., & Malik, M. K. (2019). Textual keyword extraction and summarization: State-of-the-art. *Information Processing & Management*, *56*(6), 102088. https://doi.org/10.1016/j.ipm.2019.102088

OpenAI. (2023). *GPT-4 Technical Report* (arXiv:2303.08774). arXiv. https://doi.org/10.48550/arXiv.2303.08774

Papoutsaki, A., Sangkloy, P., Laskey, J., Daskalova, N., Huang, J., & Hays, J. (2016). Webgazer: Scalable webcam eye tracking using user interactions. *Proceedings of the Twenty-Fifth International Joint Conference on Artificial Intelligence*, 3839–3845.

Patel, K., & Caragea, C. (2019). Exploring Word Embeddings in CRF-based Keyphrase Extraction from Research Papers. *Proceedings of the 10th International Conference on Knowledge Capture*, 37–44. https://doi.org/10.1145/3360901.3364447

Peng, B., Zhang, Y., Wang, M., Chen, J., & Gao, D. (2023). T-A-MFFNet: Multi-feature fusion network for EEG analysis and driving fatigue detection based on time domain network and attention network. *Computational Biology and Chemistry*, *104*, 107863. https://doi.org/10.1016/j.compbiolchem.2023.107863

Peters, M. E., Neumann, M., Iyyer, M., Gardner, M., Clark, C., Lee, K., & Zettlemoyer, L. (2018). Deep Contextualized Word Representations. *Proceedings of the 2018 Conference of the North American Chapter of the Association for Computational Linguistics: Human Language Technologies, Volume 1 (Long Papers)*, 2227–2237. https://doi.org/10.18653/v1/N18-1202

Plöchl, M., Ossandón, J. P., & König, P. (2012). Combining EEG and eye tracking: Identification, characterization, and correction of eye movement artifacts in electroencephalographic data. *Frontiers in Human Neuroscience*, *6*, 278. https://doi.org/10.3389/fnhum.2012.00278

Porzio, G. C., Rampichini, C., & Bocci, C. (Eds.). (2021). *CLADAG 2021 BOOK OF ABSTRACTS AND SHORT PAPERS: 13th Scientific Meeting of the Classification and Data Analysis Group - Firenze, September 9-11, 2021* (1st ed., Vol. 128). Firenze University Press. https://doi.org/10.36253/978-88-5518-340-6

Potnis, D., & Tahamtan, I. (2021). Hashtags for gatekeeping of information on social media. *Journal of the Association for Information Science and Technology*, *72*(10), 1234–1246. https://doi.org/10.1002/asi.24467

Proverbio, A. M., Vecchi, L., & Zani, A. (2004). From Orthography to Phonetics: ERP Measures of Grapheme-to-Phoneme Conversion Mechanisms in Reading. *Journal of Cognitive Neuroscience*, *16*(2), 301–317. https://doi.org/10.1162/089892904322984580

Puma, S., Raufaste, E., Paubel, P.-V., & El-Yagoubi, R. (2014). Fixation locked spectral analysis: Using EEG measurement in multitasking environments. *Proceedings of the International Conference on Human-Computer Interaction in Aerospace*, 1–9. https://doi.org/10.1145/2669592.2669654

Raffel, C., Shazeer, N., Roberts, A., Lee, K., Narang, S., Matena, M., Zhou, Y., Li, W., & Liu, P. J. (2023). *Exploring the Limits of Transfer Learning with a Unified Text-to-Text Transformer* (arXiv:1910.10683; Version 4). arXiv. https://doi.org/10.48550/arXiv.1910.10683

Ray Chowdhury, J., Caragea, C., & Caragea, D. (2019). Keyphrase Extraction from Disaster-related Tweets. *The World Wide Web Conference*, 1555–1566. https://doi.org/10.1145/3308558.3313696

Rodrigues, A. P., Fernandes, R., Bhandary, A., Shenoy, A. C., Shetty, A., & Anisha, M. (2021). Real-Time Twitter Trend Analysis Using Big Data Analytics and Machine Learning Techniques. *Wireless Communications and Mobile Computing*, *2021*, e3920325. https://doi.org/10.1155/2021/3920325

Sakaki, T., Okazaki, M., & Matsuo, Y. (2010). Earthquake shakes Twitter users: Real-time event detection by social sensors. *Proceedings of the 19th International Conference on World Wide Web*, 851–860. https://doi.org/10.1145/1772690.1772777

Sarrett, M. E., McMurray, B., & Kapnoula, E. C. (2020). Dynamic EEG analysis during language comprehension reveals interactive cascades between perceptual processing and sentential expectations. *Brain and Language*, *211*, 104875. https://doi.org/10.1016/j.bandl.2020.104875

Scaltritti, M., Suitner, C., & Peressotti, F. (2020). Language and motor processing in reading and typing: Insights from beta-frequency band power modulations. *Brain and Language*, *204*, 104758. https://doi.org/10.1016/j.bandl.2020.104758

Scharinger, C., Schüler, A., & Gerjets, P. (2020). Using eye-tracking and EEG to study the mental processing demands during learning of text-picture combinations. *International Journal of Psychophysiology*, *158*, 201–214. https://doi.org/10.1016/j.ijpsycho.2020.09.014

Sheoran, P., & Saini, J. S. (2020). A New Method for Automatic Electrooculogram and Eye Blink Artifacts Correction of EEG Signals using CCA and NAPCT. *Procedia Computer Science*, *167*, 1761–1770. https://doi.org/10.1016/j.procs.2020.03.386

Siu, M. H., Gish, H., Chan, A., Belfield, W., & Lowe, S. (2014). Unsupervised training of an HMM-based self-organizing unit recognizer with applications to topic classification and keyword discovery. *Computer Speech & Language*, *28*(1), 210–223. https://doi.org/10.1016/j.csl.2013.05.002

Song, M., Feng, Y., & Jing, L. (2023). A Survey on Recent Advances in Keyphrase Extraction from Pre-trained Language Models. *Findings of the Association for Computational Linguistics: EACL 2023*, 2153–2164. https://doi.org/10.18653/v1/2023.findings-eacl.161

Song, M., Jing, L., & Xiao, L. (2021). Importance Estimation from Multiple Perspectives for Keyphrase Extraction. *Proceedings of the 2021 Conference on Empirical Methods in Natural Language Processing*, 2726–2736. https://doi.org/10.18653/v1/2021.emnlp-main.215

Song, Y., Pan, S., Liu, S., Zhou, M. X., & Qian, W. (2009). Topic and keyword re-ranking for LDA-based topic modeling. *Acm Conference on Information & Knowledge Management*. https://doi.org/10.1145/1645953.1646223

Sun, Y., Qiu, H., Zheng, Y., Wang, Z., & Zhang, C. (2020). SIFRank: A New Baseline for Unsupervised Keyphrase Extraction Based on Pre-Trained Language Model. *IEEE Access*, *8*, 10896–10906. https://doi.org/10.1109/ACCESS.2020.2965087

Tarnowski, P., Kołodziej, M., Majkowski, A., & Rak, R. J. (2020). Eye-Tracking Analysis for Emotion Recognition. *Computational Intelligence and Neuroscience*, *2020*, e2909267. https://doi.org/10.1155/2020/2909267

Van Diepen, R. M., Foxe, J. J., & Mazaheri, A. (2019). The functional role of alpha-band activity in attentional processing: The current zeitgeist and future outlook. *Current Opinion in Psychology*, *29*, 229–238. https://doi.org/10.1016/j.copsyc.2019.03.015

Vashishtha, S., & Susan, S. (2021). Highlighting keyphrases using senti-scoring and fuzzy entropy for unsupervised sentiment analysis. *Expert Systems with Applications*, *169*, 114323. https://doi.org/10.1016/j.eswa.2020.114323

Vaswani, A., Shazeer, N., Parmar, N., Uszkoreit, J., Jones, L., Gomez, A. N., Kaiser, L., & Polosukhin, I. (2017). *Attention Is All You Need* (arXiv:1706.03762). arXiv. https://doi.org/10.48550/arXiv.1706.03762

Vega-Oliveros, D. A., Gomes, P. S., Milios, E. E., & Berton, L. (2019). A multi-centrality index for graph-based keyword extraction. *Information Processing & Management*, *56*(6), 102063. https://doi.org/10.1007/978-3-319-67056-0_9

Wan, X., & Xiao, J. (2008). Single document keyphrase extraction using neighborhood knowledge. *Proceedings of the 23rd National Conference on Artificial Intelligence - Volume 2*, 855–860.

Wang, J., Lou, C., Yu, R., Gao, J., Xu, T., Yu, M., & Di, H. (2018). Research on Hot Micro-blog Forecast Based on XGBOOST and Random Forest. In W. Liu, F. Giunchiglia, & B. Yang (Eds.), *Knowledge Science, Engineering and Management* (pp. 350–360). Springer International Publishing. https://doi.org/10.1007/978-3-319-99247-1_31

Wang, J., Peng, H., & Hu, J. (2006). Automatic Keyphrases Extraction from Document Using Neural Network. In D. S. Yeung, Z.-Q. Liu, X.-Z. Wang, & H. Yan (Eds.), *Advances in Machine Learning and Cybernetics* (pp. 633–641). Springer. https://doi.org/10.1007/11739685_66

Wang, X., Wei, F., Liu, X., Zhou, M., & Zhang, M. (2011). Topic sentiment analysis in twitter: A graph-based hashtag sentiment classification approach. *Proceedings of the 20th ACM International Conference on Information and Knowledge Management*, 1031–1040. https://doi.org/10.1145/2063576.2063726

Wang, X., Yu, Y., Han, F., & Wang, Q. (2022). Beta-band bursting activity in computational model of heterogeneous external globus pallidus circuits. *Communications in Nonlinear Science and Numerical Simulation*, *110*, 106388. https://doi.org/10.1016/j.cnsns.2022.106388

Wang, Y., Fan, Z., & Rose, C. (2020). Incorporating Multimodal Information in Open-Domain Web Keyphrase Extraction. *Proceedings of the 2020 Conference on Empirical Methods in Natural Language Processing (EMNLP)*, 1790–1800. https://doi.org/10.18653/v1/2020.emnlp-main.140

Wei, X., Cui, X., Cheng, N., Wang, X., Zhang, X., Huang, S., Xie, P., Xu, J., Chen, Y., Zhang, M., Jiang, Y., & Han, W. (2023). *Zero-Shot Information Extraction via Chatting with ChatGPT* (arXiv:2302.10205). arXiv. https://doi.org/10.48550/arXiv.2302.10205

Witten, I. H., Paynter, G. W., Frank, E., Gutwin, C., & Nevill-Manning, C. G. (1999). KEA: Practical automatic keyphrase extraction. *Proceedings of the Fourth ACM Conference on Digital Libraries*, 254–255. https://doi.org/10.1145/313238.313437

Won, M., Martins, B., & Raimundo, F. (2019). *Automatic extraction of relevant keyphrases for the study of issue competition* (875). Article 875. https://doi.org/10.29007/mmk4

Wu, C., Wu, F., Wu, S., Huang, Y., & Xie, X. (2018). Tweet Emoji Prediction Using Hierarchical Model with Attention. *Proceedings of the 2018 ACM International Joint Conference and 2018 International Symposium on Pervasive and Ubiquitous Computing and Wearable Computers*, 1337–1344. https://doi.org/10.1145/3267305.3274181

Xie, B., Song, J., Shao, L., Wu, S., Wei, X., Yang, B., Lin, H., Xie, J., & Su, J. (2023). From statistical methods to deep learning, automatic keyphrase prediction: A survey. *Information Processing & Management*, *60*(4), 103382. https://doi.org/10.1016/j.ipm.2023.103382

Xiong, L., Hu, C., Xiong, C., Campos, D., & Overwijk, A. (2019). *Open Domain Web Keyphrase Extraction Beyond Language Modeling* (arXiv:1911.02671). arXiv. https://doi.org/10.48550/arXiv.1911.02671

Zahera, H. M., Jalota, R., Sherif, M. A., & Ngomo, A.-C. N. (2021). I-AID: Identifying Actionable Information From Disaster-Related Tweets. *IEEE Access*, *9*, 118861–118870. https://doi.org/10.1109/ACCESS.2021.3107812

Zhang, C. (2008). *Automatic Keyword Extraction from Documents Using Conditional Random Fields* (Journal Article (Paginated) 3). Journal of Computational Information Systems; Binary Information Press. http://eprints.rclis.org/12305/

Zhang, Q., Wang, Y., Gong, Y., & Huang, X. (2016). Keyphrase Extraction Using Deep Recurrent Neural Networks on Twitter. *Proceedings of the 2016 Conference on Empirical Methods in Natural Language Processing*, 836–845. https://doi.org/10.18653/v1/D16-1080

Zhang, Y., & Zhang, C. (2019). Using Human Attention to Extract Keyphrase from Microblog Post. *Proceedings of the 57th Annual Meeting of the Association for Computational Linguistics*, 5867–5872. https://doi.org/10.18653/v1/P19-1588

Zhang, Y., & Zhang, C. (2021). Enhancing keyphrase extraction from microblogs using human reading time. *Journal of the Association for Information Science and Technology*, *72*(5), 611–626. https://doi.org/10.1002/asi.24430

Zhao, W. X., Zhou, K., Li, J., Tang, T., Wang, X., Hou, Y., Min, Y., Zhang, B., Zhang, J., Dong, Z., Du, Y., Yang, C., Chen, Y., Chen, Z., Jiang, J., Ren, R., Li, Y., Tang, X., Liu, Z., … Wen, J.-R. (2023). *A Survey of Large Language Models* (arXiv:2303.18223). arXiv. https://doi.org/10.48550/arXiv.2303.18223

Zheng, W. L., Dong, B. N., & Lu, B. L. (2014). Multimodal emotion recognition using EEG and eye tracking data. *Engineering in Medicine & Biology Society*. https://doi.org/10.1109/EMBC.2014.6944757

Zheng, W.-L., & Lu, B.-L. (2017). A multimodal approach to estimating vigilance using EEG and forehead EOG. *Journal of Neural Engineering*, *14*(2), 026017. https://doi.org/10.1088/1741-2552/aa5a98

Zheng, X., & Sun, A. (2019). Collecting event-related tweets from twitter stream. *Journal of the Association for Information Science and Technology*, *70*(2), 176–186. https://doi.org/10.1002/asi.24096

Zhu, Y., Wang, Q., & Zhang, L. (2021). Study of EEG characteristics while solving scientific problems with different mental effort. *Scientific Reports*, *11*(1), Article 1. https://doi.org/10.1038/s41598-021-03321-9